\documentclass[letterpaper]{article} 
\usepackage{aaai25}  
\usepackage{times}  
\usepackage{helvet}  
\usepackage{courier}  
\usepackage[hyphens]{url}  
\usepackage{graphicx} 
\usepackage{natbib}  
\usepackage{caption} 
\usepackage{algorithm}
\usepackage{algorithmic}
\usepackage{array}
\usepackage{multirow}
\usepackage{tabularx}
\usepackage{amsmath} 
\usepackage{amssymb}
\usepackage{booktabs} 
\usepackage{graphicx}
\usepackage{subfigure}
\usepackage{newfloat}
\usepackage{listings}
\DeclareCaptionStyle{ruled}{labelfont=normalfont,labelsep=colon,strut=off} 
\floatstyle{ruled}
\newfloat{listing}{tb}{lst}{}
\floatname{listing}{Listing}
\title{RDPI: A Refine Diffusion Probability Generation Method for Spatiotemporal Data Imputation}
\author{
    Zijin Liu\textsuperscript{\rm 1},
    Xiang Zhao\textsuperscript{\rm 2},
    You Song\textsuperscript{\rm 2}\thanks{Corresponding author.}
}
\affiliations{
    \textsuperscript{\rm 1}School of Computer Science and Engineering, Beihang University, Beijing 100191, China\\
    \textsuperscript{\rm 2}School of Software,Beihang University, Beijing 100191, China\\

    \{liuzijin, by2321118, songyou\}@buaa.edu.cn
}

\usepackage{bibentry}

\begin{document}

\maketitle

\begin{abstract}
Spatiotemporal data imputation plays a crucial role in various fields such as traffic flow monitoring, air quality assessment, and climate prediction. However, spatiotemporal data collected by sensors often suffer from temporal incompleteness, and the sparse and uneven distribution of sensors leads to missing data in the spatial dimension. Among existing methods, autoregressive approaches are prone to error accumulation, while simple conditional diffusion models fail to adequately capture the spatiotemporal relationships between observed and missing data.
To address these issues, we propose a novel two-stage \textbf{R}efined \textbf{D}iffusion \textbf{P}robability \textbf{I}mpuation (RDPI) framework based on an initial network and a conditional diffusion model. In the initial stage, deterministic imputation methods are used to generate preliminary estimates of the missing data. In the refinement stage, residuals are treated as the diffusion target, and observed values are innovatively incorporated into the forward process. This results in a conditional diffusion model better suited for spatiotemporal data imputation, bridging the gap between the preliminary estimates and the true values.
Experiments on multiple datasets demonstrate that RDPI not only achieves state-of-the-art imputation accuracy but also significantly reduces sampling computational costs. The implementation code is available at https://github.com/liuzjin/RDPI.
\end{abstract}

%

\section{Introduction}
Spatiotemporal data encompass both temporal and spatial dimensions, primarily including traffic flow data, meteorological data, environmental monitoring data, etc. playing a crucial role in scientific research and social development. However, due to equipment aging, sensor malfunctions, or communication issues, spatiotemporal data often experiences partial or even complete data loss from sensors, which severely affects the reliability and effectiveness of subsequent analyses. Therefore, effectively imputing spatiotemporal data has become a crucial research issue.

Currently, methods for addressing spatiotemporal data imputation issues are mainly categorized into deterministic methods, probabilistic methods, and diffusion methods. Deterministic methods attempt to impute values using rules or deterministic algorithms, such as Recurrent Neural Networks (RNNs) \cite{cao2018brits,  yoon2018estimating, che2018recurrent}, Transformer-based Networks 
\cite{vaswani2017attention, du2023saits, shan2023nrtsi}, and Graph Neural Networks (GNNs) \cite{cini2021filling, marisca2022learning} have been widely employed. These methods leverage the powerful representation learning capabilities of neural networks to learn complex spatiotemporal patterns from data and use learned models to predict missing data. However, deterministic methods often lack modeling of data uncertainty, which can pose challenges when dealing with complex spatiotemporal dependencies.

Probabilistic methods use statistical models or machine learning techniques to model the probability distribution of data and perform imputation based on these distributions. In addition to Variational Autoencoders (VAE)  \cite{kim2023probabilistic, mulyadi2021uncertainty, fortuin2020gp} and Generative Adversarial Networks (GAN) \cite{luo2018multivariate, liu2019naomi,  miao2021generative}, other methods such as generative models and Bayesian networks are also utilized. These methods not only learned distributions from existing data, but also leveraged the latent structures and patterns within the data to handle data uncertainty.  For example, VAE learns a latent-space representation of the data, effectively imputing missing data. 

To better simulate the diffusion process of data over time and space and capture the dynamic changes and complex dependencies of spatiotemporal data, researchers have proposed methods based on diffusion models \cite{ho2020denoising}. The Conditional Sore-based Diffusion model for Imputation (CSDI) \cite{tashiro2021csdi} is widely adopted as a typical approach. However, traditional CSDI models often only consider conditional factors during denoising network training, neglecting them in the forward and imputation process. This limitation may result in conditional models that fail to fully capture the complex dependencies in spatiotemporal traffic data, thereby affecting their performance in practical data imputation tasks.

To address the key challenges in spatiotemporal data imputation, we propose a novel two-stage refined diffusion probability imputation (RDPI) framework . In the initial stage, a preliminary estimate is generated using an initial network. In the refinement stage, a novel conditional diffusion model is proposed to optimize and adjust the results. By integrating conditional information throughout the diffusion process, the framework significantly improves the accuracy and effectiveness of imputing missing data.

Key contributions of this paper include:
\begin{itemize}
\item Introduction of the RDPI framework, which employs a two-stage imputation strategy to achieve higher-precision data completion.
\item Incorporated observed values into the forward process, deriving a conditional diffusion model better suited for spatiotemporal data imputation and utilizing residuals as the diffusion target to effectively enhance imputation efficiency.
\item Extensive empirical validation demonstrates that the proposed RDPI method excels in traffic data imputation tasks, achieving state-of-the-art performance.
\end{itemize}

These achievements provide new methodologies and theoretical support for addressing urban traffic data imputation challenges, potentially offering more reliable data foundations for urban traffic management and planning.

\section{Related Work}
\paragraph{\textbf{Deterministic Methods}}
Deterministic methods typically rely on rules or algorithms to impute missing data in spatiotemporal data. These methods learn complex spatiotemporal patterns from data and use the learned models to predict missing data. Despite their widespread application across various domains, deterministic methods often struggle to capture the complex dependencies and uncertainties present in real-world spatiotemporal data. Early research in spatiotemporal data imputation primarily relied on simple historical or global information, such as methods like KNN \cite{hastie2009elements,  beretta2016nearest} and Kriging \cite{zheng2023increase, appleby2020kriging, wu2021inductive}. With technological advancements, autoregressive methods such as ARIMA \cite{box2015time} and VAR \cite{zivot2006vector} have been introduced, which consider time periodicity and trend information for imputation. However, as data volumes grow and uncertainties in spatiotemporal  data increase, researchers have shifted towards methods that focus on extracting temporal and spatial dependencies from the data. For instance, GRAPE \cite{you2020handling} treats missing and observed data as nodes in a bipartite graph, transforming the imputation problem into a node prediction task within the graph. GRIN \cite{cini2021filling} uses message passing to reconstruct missing data across different channels, while BRITS \cite{cao2018brits} employs bidirectional recurrent neural networks to directly learn missing data within dynamic systems.
\paragraph{\textbf{Probabilistic Methods}}
Probabilistic methods utilize statistical models or machine learning techniques to model the probability distribution of spatiotemporal data and perform imputation based on these distributions. Typical probabilistic methods include VAE, GAN, and other generative models and Bayesian networks. These methods excel in handling data uncertainty by effectively imputing missing data through learning latent structures and patterns within the data. For example, GP-VAE \cite{fortuin2020gp} employs a Gaussian process model for performing variational inference, capturing correlations between features using Gaussian processes. MIWVAE \cite{mattei2019miwae}, based on Importance Weighted Autoencoders, maximizes the strict lower bound of the log-likelihood of observed data, avoiding additional costs due to missing data. SSGAN \cite{miao2021generative} introduces a novel semi-supervised adversarial network that utilizes both time and temporal label information to enhance imputation effectiveness. GAIN \cite{yoon2018gain} uses a generative adversarial approach to generate missing data by providing additional information to the discriminator through hint vectors, ensuring that the generator learns features of the real data distribution. Finally, GRUI \cite{luo2018multivariate} proposes an RNN unit that considers time lag effects and utilizes adversarial models to impute incomplete original time-series data. STGNP \cite{hu2023graph} utilizes graph neural processes and proposes a graph aggregator that accurately estimates uncertainty while capturing spatiotemporal correlations.
\paragraph{\textbf{Diffusion Methods}}
Currently, the most effective methods for addressing spatiotemporal  imputation issues are typically based on diffusion models. CSDI \cite{tashiro2021csdi} was the first time-series imputation method based on diffusion models, leveraging the Diffusion Probabilistic Models (DPMs) \cite{ho2020denoising, han2022card, li2024transformermodulated, songdenoising}to generate missing data conditioned on observed data, demonstrating the advantages of deep probabilistic generative models. SSSD \cite{alcaraz2022diffusion} integrates state-of-the-art state-space models as denoising modules and combines them with Diffwave to achieve data imputation. PriSTI \cite{liu2023pristi} extends CSDI by designing a specialized noise prediction model that extracts conditional features from enhanced observed data and calculates spatiotemporal attention weights using extracted global context priors. MIDM \cite{wang2023observed} considers a Gaussian process as the latent distribution if the end point of the diffuser in time is a standard Gaussian of missing data and observed data are paradoxes, so that the difficult priors are meaningful and easier to handle. Researchers have proposed many data imputation methods based on diffusion models, among which Conditional diffusion Data Imputation  model is widely adopted as a typical approach. However, these diffusion models often only consider conditional factors during training and denoising stages, neglecting them in the imputation process. This limitation may result in conditional models that fail to fully capture the complex dependencies in spatiotemporal traffic data, thereby affecting their performance in practical data imputation tasks.

\section{Preliminaries}
We formalize a spatiotemporal data set $\mathbf{x}\in\mathbb{R}^{N \times D} $ as sequential data consisting of N time series of length L. Any data in the dataset can be missing, meaning their values are unknown. We divide the dataset into two parts: observed data $\mathbf{x}^c$ and missing data $\mathbf{x}^m$. Specifically,  both observed and missing data are obtained from a binary mask matrix $\mathbf{m}\in \{0, 1\}^{N \times D} $, where observed data are denoted as $\mathbf{x}^c = \{\mathbf{x}^{i,j}|\mathbf{m}^{i,j}=1\}$ and missing data as $\mathbf{x}^m =\{\mathbf{x}^{i,j}|\mathbf{m}^{i,j}=0\}$.

The problem of spatiotemporal imputation can be defined as estimating $p(\mathbf{x}^m|\mathbf{x}^c)$.

\begin{figure}[h]
    \centering
    \includegraphics[width=0.47\textwidth]{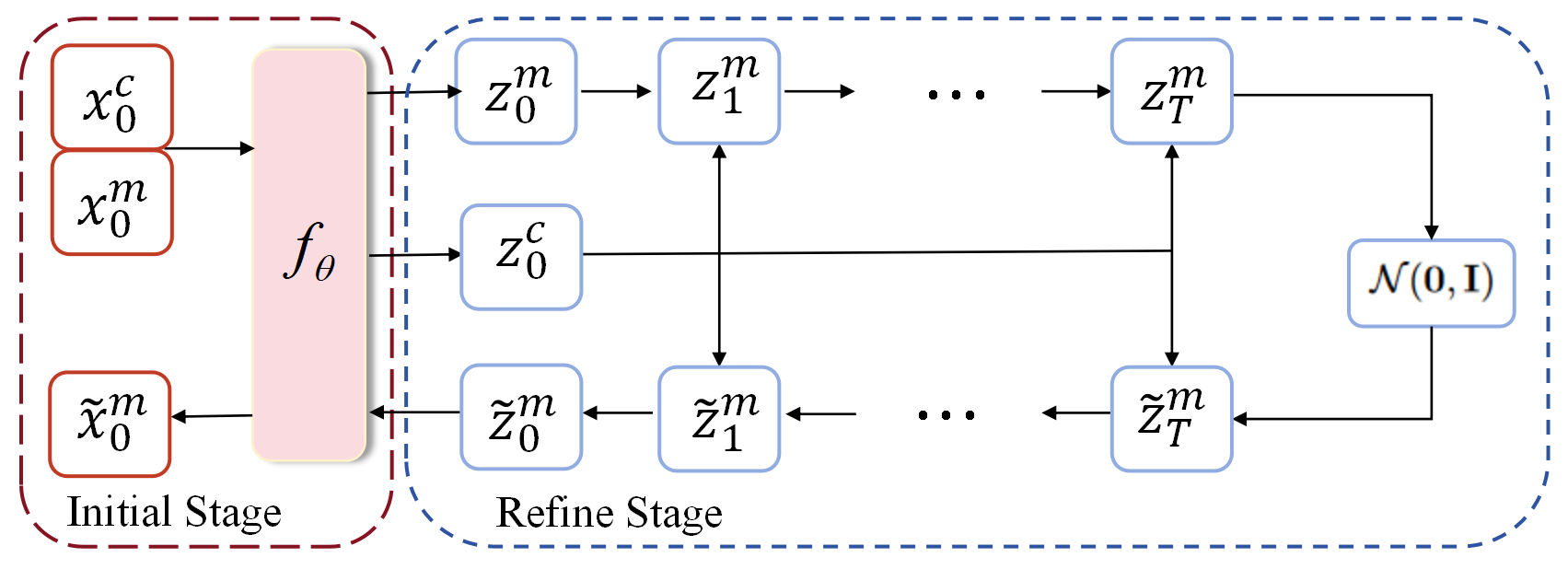}
    \caption{The directed graphical model of RDPI framework. In the initial stage, the rough imputation result $z_0^m$ is computed from the observed values $x_0^c$ and the missing data $x_0^m$ using a deterministic model $f_{\theta}$. In the refine stage, a novel conditional diffusion model is introduced to generate the residual ${z}_0^m$ between the rough imputation result $f_{\theta}(x_0^c)$ and the true values $x_0^m$, ultimately yielding a refined imputation result for the missing data. In this figure, $z_t^m$ correspond to representation of residual at $t$-th step of diffusion process, $z_0^c$ is the representation of $x_0^c$ obtained from $f_{\theta}$, $\widetilde{x}_0^m$ is the ultimate  imputation result, $\mathcal{N}(\mathbf{0}, \mathbf{I})$ correspond the Standard Gaussian. }
    \label{fig1}
\end{figure}

\section{Method}
The pipeline of our proposed Refine Diffusion Probability Imputation (RDPI) framework is illustrated in Figure \ref{fig1}. RDPI consists of two stages: initialization and refinement. In the initialization stage, initial imputation results for missing data are generated using a deterministic imputation model. Subsequently, in the refinement stage, these initial imputations are further enhanced through a novel conditional diffusion probability model. The goal of this refinement is to minimize the disparity between the initially imputed data and the true data.

Current imputation methods based on conditional diffusion models often focus solely on conditions during the reverse process, neglecting the inherent relationship between missing data and observed values in the forward process. Therefore, we rigorously defined the conditional probability distributions for the forward and backward processes, and derived a novel ELBO that incorporates the forward conditions. Based on these theoretical frameworks, we developed a novel conditional diffusion imputation model.
Thus, in the following sections, we will introduce the corresponding reverse process, forward process, the ELBO as well as the final training and imputation algorithms.

\subsection{Initial Stage}
In the proposed framework, an initial model generates preliminary missing data, which is then refined by a diffusion model that establishes a connection between these preliminary values and the true values. Probabilistic sampling of the diffusion model helps mitigate overfitting issues of the initial model on the training data while reducing the pressure of relying solely on the diffusion model for generating missing data. In theory, the initial model can employ any imputation method, including the diffusion model itself. However, using diffusion models in both stages increases randomness and the computational cost of iterative sampling. Therefore, employing a deterministic approach in the initial stage is more practical. The combination of deterministic and diffusion models preserves the rapid generation capability of deterministic models and maintains the precise likelihood calculation ability of diffusion models. Joint training in both stages further enhances the effectiveness of deterministic imputation methods. Using the deterministic model obtained through two-stage training as the final imputation model, the diffusion model can be seen as a module that improves the deterministic imputation model. Thus, our conditional diffusion model can be integrated with any deterministic missing data imputation method to enhance its imputation results.

In our experiments, we employed the deterministic method $f_\theta$ (utilizing GRIN  \cite{cini2021filling} as described in the experiment) with an initial imputation value, and the noise-disturbed object of diffusion model $g_\theta$ was the residual $\mathbf{z}_0^m$ of the true missing data $\mathbf{x}_0^m$ and initial imputation $f_\theta(\mathbf{x}_0^c)$. We utilized a two-stage training approach with joint training of the deterministic model and the diffusion model, where the training loss in the initial stage was:
\begin{equation}
\mathcal{L}_{\text{init}} =\|f_\theta({x}_0^c) -{x}_0^m \|
\end{equation}
It is worth noting that the initial model may not necessarily require pretraining, as gradients from joint training flow into  $g_\theta$ through $f_\theta$  \cite{whang2022deblurring}. However, for ensuring the stability of the diffusion model, we recommend pretraining. 
In contrast to reference \cite{whang2022deblurring}, our loss function calculates $\|f_\theta(\mathbf{x}_0^c) -\mathbf{x}_0^m \|$ rather than $\|\mathbf{x}_0^m - f_\theta(\mathbf{x}_0^c) \|$. Experimental results show that using 
$\|f_\theta(\mathbf{x}_0^c) -\mathbf{x}_0^m \|$ can increase the initial model's imputation loss if 
$f_\theta(\mathbf{x}_0^c)$ is negative, leading to instability in the diffusion target as the diffusion loss decreases. Maintaining $f_\theta(\mathbf{x}_0^c)$ positive ensures that the imputation performance of the initial model remains stable during further optimization.
\subsection{Reverse Process}
Following the steps of DDPM (Denoising Diffusion Probabilistic Models)  \cite{ho2020denoising}, the conditional diffusion model \cite{han2022card} is a latent variable model of the form $p(\mathbf{z}_0^m|\mathbf{z}_0^c) = \int p(\mathbf{z}_{0:T}^m|\mathbf{z}_0^c)d\mathbf{z}_{1:T}^m $, where $\mathbf{z}_{1:T}^m$ is a latent variable with the same dimension as the residual $\mathbf{z}_0^m$ conditioned on observed data $\mathbf{z}_0^c$ . The joint probability $p(\mathbf{z}_{0:T}^m|\mathbf{z}_0^c)$ is referred to as the reverse process, defined on a Markov chain:
\begin{equation}
\begin{aligned}
       &p(\mathbf{z}_{0:T}^m|\mathbf{z}_0^c) := p(\mathbf{z}_T^m|\mathbf{z}_0^c)\prod_{t=1}^T p(\mathbf{z}_{t-1}^m|\mathbf{z}_t^m, \mathbf{z}_0^c) \\
    &p(\mathbf{z}_{t-1}^m|\mathbf{z}_t^m, \mathbf{z}_0^c) := \mathcal{N}(\mathbf{z}_{t-1}^m;\mathbf{\mu}_{\theta}
    (\mathbf{z}_{t}^m,\mathbf{z}_0^c,t),\mathbf{\Sigma}_{\theta}(\mathbf{z}_{t}^m,\mathbf{z}_0^c,t)) 
\end{aligned}
\end{equation}
where $p(\mathbf{z}_T^m|\mathbf{z}_0^c)$ represents the endpoint of the conditional diffusion process, which is assumed to follow a standard Gaussian distribution. In this context, $\mathbf{\mu}_{\theta}$ and $\mathbf{\Sigma}_{\theta}$ denote the mean and variance, respectively, and can be estimated using neural networks. We fix the $\mathbf{\Sigma}_{\theta}$ to be constant.

\subsection{Forward Process}
Given a data point  $\mathbf{z}_0^m$ sampled from the true spatiotemporal data distribution $ p(\mathbf{z}^m)$, we define the forward process or diffusion process as a Markov chain with learned Gaussian transitions that starts from the sampled data distribution:   
\begin{equation}
\begin{aligned}
    &q(\mathbf{z}_{1:T}^m|\mathbf{z}_0^m, \mathbf{z}_0^c) := \prod_{t=1}^T q(\mathbf{z}_t^m|\mathbf{z}_{t-1}^m, \mathbf{z}_0^c)\\
    &q(\mathbf{z}_t^m|\mathbf{z}_{t-1}^m, \mathbf{z}_0^c) := \mathcal{N}(\mathbf{z}_t^m;\sqrt{1-\beta_t}\mathbf{z}_{t-1}^m+\sqrt{1-\beta_t}\mathbf{z}_0^c,\beta_t\mathbf{I})
\end{aligned}
\end{equation}
where $\beta_t$ is variance from a given variance schedule $\beta_1, \cdots, \beta_T$.

At each transition step, a small amount of Gaussian noise is added to the data through Gaussian transitions. The step size is controlled by a variable schedule $\{ \beta_t\in (0,1) \}_{t=1}^T$. Let $\overline{\alpha}_t := 1-\beta_t$ and $\alpha_t = \prod_{s=1}^t \overline{\alpha}_t$, we sample $\mathbf{z}_t^m$ directly from $\mathbf{z}_0^m$  with an arbitrary timestep t:
\begin{equation}
    q(\mathbf{z}_t^m|\mathbf{z}_0^m, \mathbf{z}_0^c) = \mathcal{N}(\mathbf{z}_t^m;\sqrt{\alpha_t}(\mathbf{z}_0^m+\mathbf{z}_0^c),(1-\alpha_t)\mathbf{I})
\end{equation}

Benefiting from the reparameterization trick, the diffusion process can be implemented as: 
\begin{equation}\label{eq5}
    \mathbf{z}_t^m = \sqrt{\alpha_t}\mathbf{z}_0^m + \sqrt{\alpha_t}\mathbf{z}_0^c+ \sqrt{(1-\alpha_t)} \epsilon_t
\end{equation}
 where $\mathbf{\epsilon} \sim \mathcal{N}(\mathbf{0},\mathbf{I})$ is random noise sampled from a standard Gaussian distribution.

\subsection{Training} 
With the light of reverse process and forward process, a novel ELBO of the conditional diffusion model can be derived as follows (See Appendix\ref{sec:a1} for details):
\begin{equation}
\label{eq6}
\begin{aligned}
    &\log p(\mathbf{z}_0^m|\mathbf{z}_0^c) \\
    &= \mathbb{E}_{q(\mathbf{z}_{1:T}^m|\mathbf{z}_0^m,\mathbf{z}_0^c)}[D_{KL}(q(\mathbf{z}_T^m|\mathbf{z}_0^m,\mathbf{z}_0^c)||p(\mathbf{z}_T^m|\mathbf{z}_0^c))\\
    &+\sum_{i=2}^TD_{KL}(q(\mathbf{z}_{t-1}^m|\mathbf{z}_t^m,\mathbf{z}_0^m, \mathbf{z}_0^c)||p(\mathbf{z}_{t-1}^m|\mathbf{z}_t^m, \mathbf{z}_0^c)) \\
                        &+  \log p(\mathbf{z}_0^m|\mathbf{z}_1^m,\mathbf{z}_0^c)]
\end{aligned}
\end{equation}
Using Bayes’ rule, we have:
\begin{equation}\label{eq7}
\begin{aligned}
        &q(\mathbf{z}_{t-1}^m|\mathbf{z}_t^m,\mathbf{z}_0^m, \mathbf{z}_0^c) = exp( -\frac{1}{2}( (\frac{\overline{\alpha}_t}{\beta_t} +\frac{1}{1-\alpha_{t-1}}){\mathbf{z}_{t-1}^m}^2 \\
        &- (\frac{2\sqrt{\overline{\alpha}_t}\mathbf{z}_t^m - 2\overline{\alpha}_t\mathbf{z}_0^c}{\beta_t} + 
        \frac{2\sqrt{{\alpha_{t-1}}}\mathbf{z}_0^m+2\sqrt{{\alpha_{t-1}}}\mathbf{z}_0^c}{1-\alpha_{t-1}}){\mathbf{z}_{t-1}^m} \\
        &+ C(\mathbf{z}_t^m,\mathbf{z}_0^m,\mathbf{z}_0^c ))) \\
\end{aligned}
\end{equation}
(See Appendix \ref{sec:a2} for details) where $C(\mathbf{z}_t^m,\mathbf{z}_0^m,\mathbf{z}_0^c )$ is some function not involving $\mathbf{z}_{t-1}^m$and details are omitted.

Following the standard Gaussian density function, the mean and variance can be parameterized as follows (See Appendix \ref{sec:a3} for details):
\begin{equation}
\begin{aligned}
    \widetilde{\beta}_t = \frac{1-\alpha_{t-1}}{1-\alpha_t}\beta_t
    \end{aligned}
\end{equation}

\begin{equation}\label{eq9}
\begin{aligned}
    &\widetilde{\mathbf{\mu}}_t
    (\mathbf{z}_{t}^m,\mathbf{z}_0^c,t)
    := \frac{\sqrt{\overline{\alpha}_t}(1-\alpha_{t-1})}{1-\alpha_t}\mathbf{z}_t^m  \\
    &+\frac{\sqrt{\alpha_{t-1}}\beta_t}{1-\alpha_t}\mathbf{z}_0^m + \frac{\sqrt{{\alpha_{t-1}}}\beta_t- \overline{\alpha}_t(1-\alpha_{t-1})}{1-\alpha_t}\mathbf{z}_0^c 
\end{aligned}
\end{equation}
we can represent $\mathbf{z}_0^m = (\mathbf{z}_t^m - \sqrt{\alpha_t}\mathbf{z}_0^c - \sqrt{(1-\alpha_t)} \mathbf{\epsilon}_t)/\sqrt{\alpha_t}$ and plug it into the above equation and obtain (See Appendix \ref{sec:a4} for details):

\begin{equation}
\begin{aligned}
&\widetilde{\mathbf{\mu}}_t(\mathbf{z}_{t}^m,\mathbf{z}_0^c,t)\\
    &= \frac{1}{\sqrt{\overline{\alpha}_t}}(\mathbf{z}_t^m  -\frac{\overline{\alpha}_t\sqrt{\overline{\alpha}_t}(1-\alpha_{t-1})}{1-\alpha_t}\mathbf{z}_0^c-\frac{1-\overline{\alpha}_t}{\sqrt{(1-\alpha_t) }}\mathbf{\epsilon}_t)\\
\end{aligned}
\end{equation}

We need to learn a neural network to estimate the conditioned probability distributions in the reverse diffusion process $p(\mathbf{z}_{t-1}^m|\mathbf{z}_t^m, \mathbf{z}_0^c) = \mathcal{N}(\mathbf{z}_{t-1}^m;\mathbf{\mu}_{\theta}
    (\mathbf{z}_{t}^m,\mathbf{z}_0^c,t),\mathbf{\Sigma}_{\theta}(\mathbf{z}_{t}^m,\mathbf{z}_0^c,t))$. We would like to train $\mathbf{\mu}_{\theta}$  to predict $\widetilde{\mathbf{\mu}}_t$. We can reparameterize the Gaussian noise term instead to make it predict $\epsilon$ from the input $\mathbf{z}_{t}^m$ at time step $t$:
\begin{equation}\label{sample}
\begin{aligned}
    \mathbf{\mu}_{\theta}(\mathbf{z}_t^m,\mathbf{z}_0^c, t)&= \frac{1}{\sqrt{\overline{\alpha}_t}}(\mathbf{z}_t^m -\frac{\overline{\alpha}_t\sqrt{\overline{\alpha}_t}(1-\alpha_{t-1})}{1-\alpha_t}\mathbf{z}_0^c \\
    &-\frac{1-\overline{\alpha}_t}{\sqrt{(1-\alpha_t) }}\mathbf{\epsilon}_{\theta}(\mathbf{z}_t^m, \mathbf{z}_0^c, \mathbf{z}_0^c,t))
\end{aligned}
\end{equation}
 The loss term can be written (See Appendix \ref{sec:a5} for details):
\begin{equation}
\begin{aligned}
    \mathcal{L}_t = \mathbb{E}_{\mathbf{z}_0^m, \epsilon}[\frac{(1-\overline{\alpha}_t)^2}{2(1-\alpha_t)\|\mathbf{\Sigma}_{\theta} \|_2^2}\|\mathbf{\epsilon}_t- \mathbf{\epsilon}_{\theta}(\mathbf{z}_t^m, \mathbf{z}_0^c,\mathbf{z}_0^c, t)\|]\\
\end{aligned}
\end{equation}
The Simplified training objective can be written: 
\begin{equation}\label{loss}
\begin{aligned}
    \mathcal{L}_{\text{simple}} &= \mathbb{E}_{\mathbf{z}_0^m, \epsilon}[\|\mathbf{\epsilon}_t- \mathbf{\epsilon}_{\theta}(\mathbf{z}_t^m, \mathbf{z}_0^c,\mathbf{z}_0^c, t)\|]\\
\end{aligned}
\end{equation}
We follow the convention to assume it will be close to zero by carefully diffusing the observed response variable $\mathbf{z}_0^c$ towards a pre-assumed distribution $p(\mathbf{z}_T^m|\mathbf{z}_0^c)$.

At last, joint loss of RDPI becomes:
\begin{equation}
    \mathcal{L}_{\text{joint}} = \mathcal{L}_{\text{simple}} + \lambda\mathcal{L}_{\text{init}}
\end{equation}
where $\lambda$ is the hyperparameter that trades off between the initial loss and diffusion loss.
The complete training procedure has been displayed in Algorithm \ref{alg1}.

\begin{algorithm}
    \caption{Training of RDPI}
    \label{alg1}
    \begin{algorithmic}[1]
        \STATE Pre-train the deterministic imputation model $f_\theta$
        \WHILE{not converged}
            \STATE Draw $\mathbf{x}_0^m \sim q(\mathbf{x}_0^m|\mathbf{x}_0^c)$
            \STATE $\mathbf{z}_0^m =  f_\theta(\mathbf{x}_0^c)-\mathbf{x}_0^m $
            \STATE Compute initial loss $\mathcal{L}_{\text{init}} = \|\mathbf{z}_0^m\|$
            \STATE Draw $t \sim \text{Uniform}(\{1,\cdots,T\})$
            \STATE Draw $\boldsymbol{\epsilon} \sim \mathcal{N}(\mathbf{0}, \mathbf{I})$
            \STATE Compute diffusion loss \\$\mathcal{L}_{\text{simple}} = \|\boldsymbol{\epsilon}_t - \boldsymbol{\epsilon}_{\theta}(\mathbf{z}_t^m, \mathbf{z}_0^c, \mathbf{z}_0^c, t)\|^2$
            \STATE Compute joint loss $\mathcal{L}_{\text{joint}} = \mathcal{L}_{\text{simple}} + \lambda\mathcal{L}_{\text{init}}$
            \STATE Take gradient descent step on $\nabla_\theta\mathcal{L}_{\text{joint}}$
        \ENDWHILE
    \end{algorithmic}
\end{algorithm}

\begin{algorithm}
    \caption{Imputating (Sampling) with RDPI}
    \label{alg2}
    \begin{algorithmic}[1]
        \STATE $\mathbf{x}_{init}^m = f_\theta(\mathbf{x}_0^c)$
        \STATE $\mathbf{z}_T^m \sim \mathcal{N}(\mathbf{0}, \mathbf{I})$
        
        \FOR{$t = T$ to $1$}
            \STATE $\boldsymbol{\epsilon}_t \sim \mathcal{N}(\mathbf{0}, \mathbf{I})$ if {$t > 1$} else $\boldsymbol{\epsilon}_t = \mathbf{0}$ 
            \STATE $\mathbf{z}_{t-1}^m = \frac{1}{\sqrt{\overline{\alpha}_t}} \Bigg( \mathbf{z}_t^m - \frac{\overline{\alpha}_t\sqrt{\overline{\alpha}_t}(1-\alpha_{t-1})}{1-\alpha_t}\mathbf{z}_0^c$
            \STATE \hspace{4em}$- \frac{1-\overline{\alpha}_t}{\sqrt{(1-\alpha_t)}} \mathbf{\epsilon}_{\theta}(\mathbf{z}_t^m, \mathbf{z}_0^c, \mathbf{z}_0^c, t) \Bigg) + \sigma_t \boldsymbol{\epsilon}_t$
        \ENDFOR
        \RETURN $\mathbf{x}_{init}^m - \mathbf{z}_0^m$
    \end{algorithmic}
\end{algorithm}
    
\subsection{Imputation}
After training denoising networks in each diffusion step, we can derive the likelihood of residual distributions according to the Eq \eqref{sample}. Missing data can be generated by sampling from standard Gaussian noise, as outlined in Algorithm \ref{alg2}. Similarly to DDPM, the significant computational demand of iterative sampling is a primary drawback of diffusion models. To expedite sampling, RDPI can employ any unconditional DDPM accelerated sampling scheme, such as DDIM  \cite{songdenoising}. However, it's crucial to note that unlike the original DDIM, a version without randomness may compromise the quality of final data imputation. Therefore, RDPI values the specific sampling approach by retaining the accelerated scheme with intact random components.

Following the steps of DDIM, we can get 
\begin{equation}
\begin{aligned}
    p(\mathbf{z}_{t-1}^m|\mathbf{z}_t^m, \mathbf{z}_0^c) &= \mathcal{N}(\mathbf{z}_{t-1}^m;\sqrt{\frac{\alpha_{t-1}}{\alpha_{t}}}\mathbf{z}_t^m+\\
    &(\sqrt{1-\alpha_{t-1}-\widetilde{\beta_t}^2} -\sqrt{\frac{\alpha_{t-1}(1-\alpha_t)}{\alpha_t}})\epsilon_{\theta}, \widetilde{\beta_t})
\end{aligned}
\end{equation}
(See Appendix \ref{subsec:a6} for details). 
The pipeline of the accerlerate imputation method has been displayed in  Algorithm \ref{alg3}.

    

\begin{algorithm}
    \caption{Accelerated Imputating}
    \label{alg3}
    \begin{algorithmic}[1]
        \STATE $x_{init}^m = f_\theta(\mathbf{x}_0^c)$
        \STATE $\mathbf{z}_T^m \sim \mathcal{N}(\mathbf{0}, \mathbf{I})$
        
        \FOR{$t = T$ to $1$}
            \STATE $\boldsymbol{\epsilon}_t \sim \mathcal{N}(\mathbf{0}, \mathbf{I})$ if {$t > 1$} else $\boldsymbol{\epsilon}_t = \mathbf{0}$ 
            \STATE $\mathbf{z}_{t-1}^m = \sqrt{\frac{\alpha_{t-1}}{\alpha_{t}}}\mathbf{z}_t^m+(\sqrt{1-\alpha_{t-1}-\widetilde{\beta_t}^2} -\sqrt{\frac{\alpha_{t-1}(1-\alpha_t)}{\alpha_t}})\epsilon_{\theta} + \widetilde{\beta_t}\boldsymbol{\epsilon}_t$
        \ENDFOR
        \RETURN $\mathbf{x}_{init}^m - \mathbf{z}_0^m$
    \end{algorithmic}
\end{algorithm}

\subsection{Denoising Model}
\begin{figure}[h]
    \centering
    \includegraphics[width=0.45\textwidth]{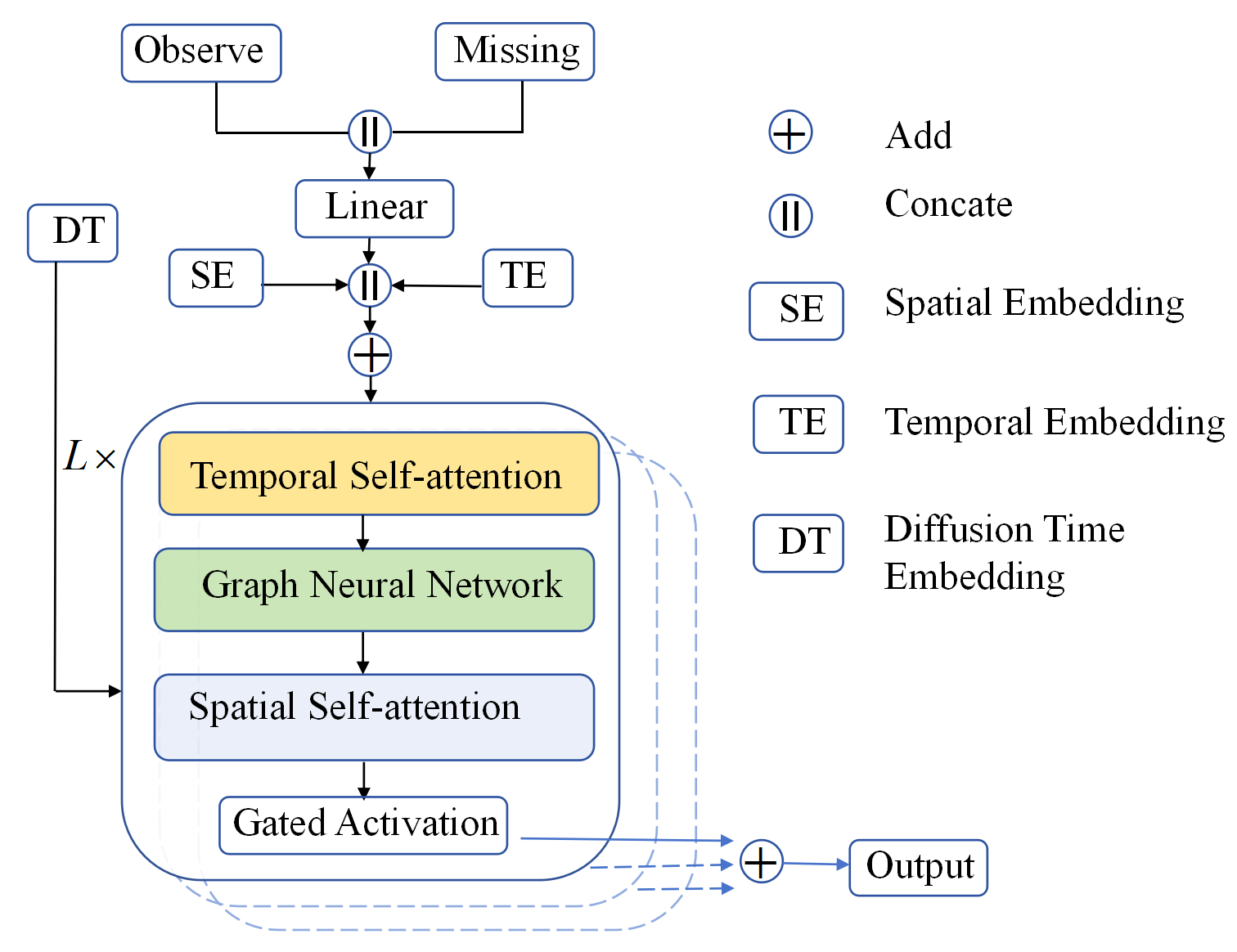}
    \caption{Architecture of denoising model.}
    \label{fig2}
\end{figure}

In this section, we introduce the proposed denoising model. The specific process is illustrated in Figure \ref{fig2}. Our denoising model consists of four components: embedding module,  temporal self-attention module, graph neural network module and  spatial self-attention module.

To keep the native information in the raw data, we first concatenates observed and missing data, then utilize a 1D convolutional layer to obtain a feature embedding $E^f \in \mathbb{R}^{T \times N\times d}$:
\begin{equation}
    {E}^f = \text{CNN}(\mathbf{z}_0^c ||\mathbf{z}_0^m)
\end{equation}

In spatiotemporal sequences, data from the same node exhibit temporal heterogeneity over time, whereas data from different nodes at the same time point show spatial heterogeneity across space. Therefore, it is essential to encode the time information and spatial features into the data. The embedding module is responsible for converting time information, spatial information, and diffusion step information into trainable embedding tables. Traffic flow data exhibit distinct periodic patterns, with weekdays and weekends showing temporal heterogeneity within the same week, and significant variations occurring throughout the day. Thus, we encode the time window into a learnable embedding table $E_d^{tem} \in \mathbb R ^{N_W \times d}$,  where $N_W$ and $d$ is the lenth of time window and the dimension of the feature embedding. Furthermore, we designed a spatial embedding $E_d^{spa} \in \mathbb R ^{N_S \times d}$ to simultaneously capture temporal and spatial heterogeneity, where $N_S$ and $d$ denote  the number of nodes and  the dimension of the feature embedding, respectively. The final representation of the spatiotemporal embedding $Z^{emb}$ is obtained by summing up all the embeddings:
\begin{equation}
    Z^{emb} =E^f+E_d^{tem}+E_d^{spa}
\end{equation}

Since the attention mechanism performs exceptionally well in modeling sequential data, we use self-attention to model the spatial and temporal dependencies of $Z_t^m$, which can be formulated as follows:
\begin{equation}
\begin{aligned}
&\text{Attn}(Z^{emb}) = \mathrm{softmax}\left( \frac{QK^T}{\sqrt{d}} \right)V \ \\
& Q = W_Q \cdot Z^{emb}, \quad K = W_K \cdot Z^{emb}, \quad V = W_V \cdot Z^{emb}
\end{aligned}
\end{equation}
where $W_Q$, $W_K$ and $W_V \in \mathbb{R}^{d \times d}$ are learnable projection parameters.
To capture the relationships between nodes in spatiotemporal data, we employ graph neural networks to further model spatial dependencies which can be  formalized as follows:
\begin{equation}
     Z^{emb}= \text{GNN}(Z^{emb}, A)
\end{equation}
where $A$ denotes adjacency matrix.
\section{Experiments}
In this section, we first introduce the experimental dataset, followed by the comparison methods, evaluation metrics, and experimental settings. We evaluated the proposed RDPI framework through extensive spatiotemporal imputation experiments, aiming to answer the following questions:

RQ 1: Can RDPI demonstrate stronger imputation capabilities compared to state-of-the-art methods?

RQ 2: How does the imputation performance of RDPI vary under different missing-rate conditions?

RQ 3: Does the imputation of missing data benefit from conditional observational data and the initial imputation model?

RQ 4: Does RDPI exhibit good imputation performance even when the entire node is missing?

RQ 5: Does hyperparameter tuning significantly impact on the model's performance?

\subsection{Datasets and Experimental Settings}
\paragraph{\textbf{Datasets.}}We conducted comparative experiments on four real datasets: PEMS-BAY  \cite{li2017diffusion}, METR-LA  \cite{li2017diffusion}, AQI  \cite{yi2016st}and AQI36  \cite{yi2016st}. PEMS-BAY contains traffic speed data from 207 nodes on California highways over four months. METR-LA collected traffic speed data from 325 nodes on Los Angeles highways over six months. Both PEMS-BAY and METR-LA are sampled every 5 minutes. AQI contains hourly sampled air quality data from 437 observation nodes in 43 Chinese cities over twelve months. AQI36 contains 36 observation nodes from AQI in Beijing over twelve months.  Table \ref{tab1} shows the number of nodes, time slices, and missing data situations for three datasets.

\begin{table}[h]
    \centering
    \begin{tabular}{cccc}
    \toprule
        Dataset & nodes & Time Slices & Missing rate(\%)  \\
     \midrule
         PEMS-BAY &  325 &  52128 &  0.023\\
         METR-LA &  207 & 34272 & 8.11\\
         AQI36 & 36 & 8759 & 13.24\\
         AQI & 437 & 8760 & 25.67\\
      \bottomrule
    \end{tabular}
    \caption{Statistics of datasets.}
    \label{tab1}
\end{table}
\paragraph{\textbf{Settings.}} For fairness, as done in previous work  \cite{cini2021filling, liu2023pristi}, we split PEMS-BAY and METR-LA into training, validation, and test sets in a ratio of 70\%/10\%/20\%. For AQI and AQI36, the 3th, 6th, 9th and 12th months are used as test sets and the other months as the training sets. The hyperparameters of all data sets were summaried in Table \ref{tab2}. All the experiments are run for 5 times.
\begin{table}
    \centering
    \begin{tabular}{ccccc}
    \toprule
        Description& BAY& LA& AQI36 & AQI  \\
     \midrule
         Batch size &  128 &  128 &  128 & 128\\
         Time length &  24 & 24 & 36 &36\\
         Epochs & 300 & 300 & 200 &200\\
         Learning rate & 1e-3 & 1e-3 & 1e-3& 1e-3\\
         Minimum noise level & 1e-4  & 1e-4 & 1e-4& 1e-4\\
         Maximum noise level & 0.2  & 0.02 & 0.02& 0.02\\
         Diffusion steps& 50  & 50 & 100& 100\\
         Accelerate steps& 10  & 10 & 40& 40\\
         Loss balancing parameter &0.2&0.2&0.5&0.5\\
      \bottomrule
    \end{tabular}
    \caption{The hyperparameters for all datasets.}
    \label{tab2}
\end{table}

\paragraph{\textbf{Training Strategies.}}In air quality datasets, missing data and their corresponding true data are already annotated in the dataset. As in previous works  \cite{cini2021filling, liu2023pristi}, we employed two different training scenarios: (1) \textbf{In-sample}, where model was trained according to the specified missing positions in the dataset; (2) \textbf{Out-of-sample}, where the training set combined the missing data specified with the observed data, and new missing positions were randomly generated. In traffic datasets, missing data is sparse and lacks ground truth values. Therefore, in addition to the actual observed data, we employed a method to manually design target data imputation using random masking. Similar to prior studies, the imputation targets were divided into two scenarios: (1) \textbf{Point missing}, where 25\% of observed data was randomly masked; (2) \textbf{Block missing}, where initially 5\% of observed data was masked, and then for each node, data within 1 to 4 hours was randomly masked at a probability of 0.15\%.
\begin{table*}
  \centering
    \begin{tabular}{clcccccc}
    \toprule
     \multirow{2}[3]{*}{\centering Dataset} & \multirow{2}[3]{*}{\centering Model}& \multicolumn{3}{|c}{In-sample}& \multicolumn{3}{c}{Out-of-sample } \\
     \cmidrule{3-8} 
      & &\multicolumn{1}{|c}{MAE} & MSE& MRE(\%)  &\multicolumn{1}{|c}{MAE} &MSE &MRE(\%) \\
    \midrule
    
    \multirow{11}{*}{\centering \rotatebox{90}{AQI-36}}
         &\multicolumn{1}{|c|}{MEAN} &  53.48$\pm$0.00& 4578.08$\pm$00.00 &76.77$\pm$0.00  &\multicolumn{1}{|c}{53.48$\pm$0.00}& 4578.08$\pm$00.00& 76.77$\pm$0.00 \\
     
     &\multicolumn{1}{|c|}{KNN}  &  30.21$\pm$0.00&2892.31$\pm$00.00& 43.36$\pm$0.00 & \multicolumn{1}{|c}{30.21$\pm$0.00}& 2892.21$\pm$00.00&43.36$\pm$0.00\\

     &\multicolumn{1}{|c|}{MICE} &  29.89$\pm$0.11&2575.53$\pm$07.67 & 42.90$\pm$0.15& \multicolumn{1}{|c}{30.37$\pm$0.09}& 2594.06 $\pm$07.17&43.59$\pm$0.13\\

     &\multicolumn{1}{|c|}{VAR} &  13.16$\pm$0.21&513.90$\pm$12.39& 18.89$\pm$0.31&\multicolumn{1}{|c}{ 15.64$\pm$0.08}& 833.46$\pm$13.85& 22.02$\pm$0.11\\

     &\multicolumn{1}{|c|}{BRITS}&  12.24$\pm$0.26&495.94$\pm$43.56 & 17.57$\pm$0.38&\multicolumn{1}{|c}{ 14.50$\pm$0.35}&662.36$\pm$65.16&20.41$\pm$0.50\\
     
     &\multicolumn{1}{|c|}{GRIN}&  10.51$\pm$0.28&371.47$\pm$17.38 & 15.09$\pm$0.40&\multicolumn{1}{|c}{12.08$\pm$0.47}&523.14$\pm$57.17&17.00$\pm$0.67\\

     &\multicolumn{1}{|c|}{rGAIN}&  12.23$\pm$0.17&393.76$\pm$12.66 & 17.55$\pm$0.25& \multicolumn{1}{|c}{15.37$\pm$0.26}&641.92$\pm$33.89&21.63$\pm$0.36\\
     
     &\multicolumn{1}{|c|}{GP-VAE}& 14.11$\pm$0.24&483.91$\pm$24.36 & 18.43$\pm$0.45& \multicolumn{1}{|c}{25.71$\pm$0.30 }& 2589.53$\pm$ 59.14 & -\\
     
     &\multicolumn{1}{|c|}{CSDI}&  9.60$\pm$0.14&372.49$\pm$16.90&  15.49$\pm$0.37& \multicolumn{1}{|c}{9.51$\pm$0.10}& 352.46$\pm$7.50 &-\\
     \cmidrule{2-8}
     &\multicolumn{1}{|c|}{MIDM}&  9.41$\pm$0.20&361.28$\pm$21.33 & 14.87$\pm$0.41& \multicolumn{1}{|c}{-}& - & -\\

     &\multicolumn{1}{|c|}{RDPI }&  \textbf{7.98$\pm$0.24}  &   \textbf{238.25$\pm$13.22}&  \textbf{11.67$\pm$0.35 }&  \multicolumn{1}{|c}{\textbf{9.14$\pm$0.03}}& \textbf{306.40$\pm$21.33} &  \textbf{13.45 $\pm$0.02}\\
     \midrule
     \multirow{11}{*}{\centering \rotatebox{90}{AQI}}
    
      &\multicolumn{1}{|c|}{MEAN} &  39.60$\pm$0.00 & 3231.04$\pm$00.00 &59.25$\pm$0.00 
      &\multicolumn{1}{|c}{39.60$\pm$0.00} & 3231.04$\pm$00.00 & 59.25$\pm$0.00  \\
     
     &\multicolumn{1}{|c|}{KNN}  & 34.10$\pm$0.00 & 3471.14$\pm$00.00 & 51.02$\pm$0.00 
     &\multicolumn{1}{|c}{34.10$\pm$0.00} & 3471.14$\pm$00.00 & 51.02$\pm$0.00 	 \\

     &\multicolumn{1}{|c|}{MICE} & 26.39$\pm$0.13 & 1872.53$\pm$15.97& 39.49$\pm$0.19 
     &\multicolumn{1}{|c}{26.98$\pm$0.10} & 1930.92$\pm$10.08 & 40.37$\pm$0.15\\

     &\multicolumn{1}{|c|}{VAR} & 18.13$\pm$0.84 & 918.68$\pm$56.55 & 27.13$\pm$1.26 
     &\multicolumn{1}{|c}{22.95$\pm$0.30} & 1402.84$\pm$52.63 & 33.99$\pm$0.44\\

     &\multicolumn{1}{|c|}{BRITS}& 17.24$\pm$0.13 & 924.34$\pm$18.26 & 25.79$\pm$0.20
     &\multicolumn{1}{|c}{20.21$\pm$0.22} &1157.89$\pm$25.66 & 29.94$\pm$0.33\\
     
     &\multicolumn{1}{|c|}{GRIN}& 13.10$\pm$0.08 & 615.80$\pm$10.09 & 19.60$\pm$0.11 
     &\multicolumn{1}{|c}{14.73$\pm$0.15} & 775.91$\pm$28.49 & 21.82$\pm$0.23\\  
     
     &\multicolumn{1}{|c|}{rGAIN}& 17.69$\pm$0.17 & 861.66$\pm$17.49 & 26.48$\pm$0.25
     &\multicolumn{1}{|c}{21.78$\pm$0.50} &1274.93$\pm$60.28 & 32.26$\pm$0.75\\
     
     &\multicolumn{1}{|c|}{GP-VAE}&17.84$\pm$0.16 &893.27$\pm$20.39 &27.46$\pm$0.19
     &\multicolumn{1}{|c}{-} &-  &-\\
     
     &\multicolumn{1}{|c|}{CSDI}& 11.37$\pm$0.12 &589.31$\pm$11.20 & 18.26$\pm$0.24
     &\multicolumn{1}{|c}{-} &-   &- \\
     \cmidrule{2-8}
     &\multicolumn{1}{|c|}{MIDM}& 10.06$\pm$0.11 &562.84$\pm$12.01 & \textbf{16.87$\pm$0.19}
     &\multicolumn{1}{|c}{-} &-   &- \\

     &\multicolumn{1}{|c|}{RDPI}&  \textbf{9.10$\pm$0.33}&\textbf{266.81$\pm$13.65} & 17.17$\pm$0.15 &  \multicolumn{1}{|c}{\textbf{11.22$\pm$0.21}}&\textbf{388.63$\pm$0.22} & \textbf{21.16$\pm$0.02}\\
     \bottomrule[1pt]
    \end{tabular}
        \caption{Imputation results on AQI-36 and AQI. Performance averaged over 5 runs.}\label{tab3}%
\end{table*}

\subsection{Baseline}
In our study, we conducted a comparative analysis of various baseline methods,  include: (1) \textbf{MEAN}, imputation using the node-level average; (2) \textbf{KNN}, imputation by averaging values of $k=10$ highest weight neighboring nodes; (3) \textbf{MICE}  \cite{white2011multiple},  with 100 maximum iterations and 10 nearest features limitation; (4) \textbf{VAR}, a vector autoregressive single-step predictor; (5) \textbf{GRIN}  \cite{cini2021filling},  a bidirectional GRU based framework with graph neural network for multivariate time series; (6) \textbf{BRITS}  \cite{cao2018brits} using recurrent dynamics to impute the missing data; (7) \textbf{rGAIN}, GAIN with a bidirectional recurrent encoder and decoder; (8) \textbf{GP-VAE} \cite{fortuin2020gp}, a deep probabilistic model by combining VAE and Gaussian process; (9) \textbf{CSDI}  \cite{tashiro2021csdi}, a probability imputation method based on conditional diffusion models;  (10) \textbf{MIDM} \cite{wang2023observed}, a diffusion based model with noise sampling and denoising mechanism for multivariate time series imputation.

\subsection{Metrics}
    We apply three evaluation metrics to measure the performance of spatiotemporal imputation: Mean Absolute Error (MAE), Mean Squared Error (MSE), and Mean relative error (MRE), which are defined as:
    \begin{equation}
    \begin{aligned}
        &MAE(X,\hat{X})=\mathrm{mean}(\mathrm{sum}(|X-\hat{X}|)) \\
        &MSE(X,\hat{X})=\mathrm{mean}(\mathrm{sum}((X-\hat{X})^2)) \\
        &MRE(X,\hat{X})=\mathrm{mean}(\mathrm{sum}(|\frac{X-\hat{X}}{X}|))
    \end{aligned}
    \end{equation}

\subsection{Results (RQ1)}
We first evaluate the spatiotemporal imputation performance of RDPI compared with other baselines. Table \ref{tab3} presents the imputation results on the AQI36 and AQI datasets under both In-sample and Out-of-sample modes. Table \ref{tab4} presents the results of the imputation of missing data in the PEMS-BAY and METR-LA datasets under different missing patterns. Compared to all baseline methods, RDPI exhibits the best imputation performance on all three metrics. Specifically, in the In-sample mode, RDPI reduces the MSE performance relative to the nearest baseline by more than 34\% in AQI36 and by more than 50\% in AQI.  Furthermore, RDPI significantly improves performance compared to the best baseline method based on diffusion models.  Despite the relatively sparse spatial correlation between nodes in traffic scenarios, RDPI still achieves the best performance in three metrics for PEMS-BAY and METR-LA datasets, indicating its stronger generalizability in handling varying spatial dependencies.

In Table \ref{tab3} and Table \ref{tab4}, the MAE and MSE performance consistently shows the best performance, while the  MRE performance does not achieve the same level of effectiveness. We contend that the MRE may not fully capture the true performance of the model. This discrepancy arises because the diffusion model smooths the prediction residuals through probabilistic sampling, thereby reducing the magnitude of residuals between the true values and the initial imputed values. Since MSE evaluates the overall discrepancy between predicted and true values, this smoothing effect significantly improves it. In contrast, MRE focuses on relative error and may not adequately reflect the improvements resulting from this smoothing.
\begin{table*}
  \centering
    \begin{tabular}{clcccccc}
     \toprule[1pt]
     \multirow{2}[3]{*}{\centering Dataset} & \multirow{2}[3]{*}{\centering Model}& \multicolumn{3}{|c}{Block-missing}& \multicolumn{3}{c}{Point-missing } \\
     \cmidrule{3-8} 
      & &\multicolumn{1}{|c}{MAE} & MSE& MRE(\%)  &\multicolumn{1}{|c}{MAE} &MSE &MRE(\%) \\
    \midrule
    
    \multirow{11}{*}{\centering \rotatebox{90}{PEMS-BAY}}
         &\multicolumn{1}{|c|}{MEAN} &  5.46$\pm$0.00& 87.56$\pm$ 0.00&8.75$\pm$0.00  &\multicolumn{1}{|c}{5.42$\pm$0.00}& 86.59$\pm$0.00& 8.67$\pm$0.00 \\
     
     &\multicolumn{1}{|c|}{KNN}  &  4.30$\pm$0.00&49.90$\pm$0.00& 6.90$\pm$0.00 & \multicolumn{1}{|c}{4.30$\pm$0.00}& 49.80$\pm$0.00&6.88$\pm$0.00\\

     &\multicolumn{1}{|c|}{MICE} &  2.94$\pm$0.02&28.28$\pm$0.37 & 4.71$\pm$0.03& \multicolumn{1}{|c}{3.09$\pm$0.02}& 31.43 $\pm$0.41&4.95$\pm$0.02\\

     &\multicolumn{1}{|c|}{VAR} &  2.09$\pm$0.10&16.06$\pm$0.73& 3.35$\pm$0.02&\multicolumn{1}{|c}{ 1.30$\pm$0.00}& 6.52$\pm$0.01& 2.07$\pm$0.01\\ 
     
     &\multicolumn{1}{|c|}{BRITS}&  1.70$\pm$0.01&10.50$\pm$0.07 & 2.72$\pm$0.01&\multicolumn{1}{|c}{ 1.47$\pm$0.00}&7.94$\pm$0.03&2.36$\pm$0.00\\
     
     &\multicolumn{1}{|c|}{GRIN}&  1.14$\pm$0.01&6.60$\pm$0.10 & 1.83$\pm$0.02&\multicolumn{1}{|c}{ 0.67$\pm$0.00}&1.55$\pm$0.01&1.08$\pm$0.00\\  
     
     &\multicolumn{1}{|c|}{rGAIN}&  2.18$\pm$0.01&13.96$\pm$0.20 & 3.50$\pm$0.02& \multicolumn{1}{|c}{1.88$\pm$0.02}&10.37$\pm$0.20&3.01$\pm$0.04\\
     
     &\multicolumn{1}{|c|}{GP-VAE}&  2.39$\pm$0.03&14.81$\pm$0.15 & 4.32$\pm$0.02& \multicolumn{1}{|c}{1.92$\pm$0.01}&12.43$\pm$0.08&3.67$\pm$0.02\\
     
     &\multicolumn{1}{|c|}{CSDI}&  1.16$\pm$0.01&7.02$\pm$0.09&  1.96$\pm$0.01& \multicolumn{1}{|c}{0.83$\pm$0.00}&1.79$\pm$0.00&1.42$\pm$0.00\\
     \cmidrule{2-8}
     &\multicolumn{1}{|c|}{MIDM}&  1.03$\pm$0.01&5.83$\pm$0.11 & 1.77$\pm$0.02& \multicolumn{1}{|c}{0.60$\pm$0.00}&1.54$\pm$0.02&0.93$\pm$0.00\\

     &\multicolumn{1}{|c|}{RDPI }&  \textbf{0.90$\pm$0.01}  &   \textbf{4.76$\pm$0.02}&  \textbf{1.45 $\pm$0.01 }&  \multicolumn{1}{|c}{\textbf{0.59$\pm$0.02}}& \textbf{1.42$\pm$0.01} &  \textbf{0.90$\pm$0.02}\\
     \midrule
     \multirow{11}{*}{\centering \rotatebox{90}{METR-LA}}
    
      &\multicolumn{1}{|c|}{MEAN} &  7.48$\pm$0.00&139.54$\pm$0.00 &12.96$\pm$0.00 &
      \multicolumn{1}{|c} {7.56$\pm$0.00} &142.22$\pm$0.00&13.10$\pm$0.00  \\
     
     &\multicolumn{1}{|c|}{KNN}  & 7.79$\pm$0.00 &124.61$\pm$0.00 &13.49$\pm$0.00 &
     \multicolumn{1}{|c} {7.88$\pm$0.00} &129.29$\pm$0.00&13.65$\pm$0.00 	 \\

     &\multicolumn{1}{|c|}{MICE} & 4.22$\pm$0.05 & 51.07$\pm$1.25& 7.31$\pm$0.09 &
     \multicolumn{1}{|c} {4.42$\pm$0.07} &55.07$\pm$1.46&7.65$\pm$0.12\\

     &\multicolumn{1}{|c|}{VAR} & 3.11$\pm$0.08 &28.00$\pm$0.76 &5.38$\pm$0.13 &
     \multicolumn{1}{|c} {2.69$\pm$0.00} &21.10$\pm$0.02&4.66$\pm$0.00\\                  
     
     &\multicolumn{1}{|c|}{BRITS}& 2.34$\pm$0.01 & 17.00$\pm$0.14&4.05$\pm$0.01 &
     \multicolumn{1}{|c} {2.34$\pm$0.00} &16.46$\pm$0.05&4.05$\pm$0.00\\
     
     &\multicolumn{1}{|c|}{GRIN}& 2.03$\pm$0.00 &13.26$\pm$0.05 &3.52$\pm$0.01 &
     \multicolumn{1}{|c} {1.91$\pm$0.00} &10.41$\pm$0.003&3.30$\pm$0.00\\     
     
     &\multicolumn{1}{|c|}{rGAIN}& 2.90$\pm$0.01&21.67$\pm$0.15 &5.02$\pm$0.02 &
     \multicolumn{1}{|c} {2.83$\pm$0.01} &20.03$\pm$0.09&4.91$\pm$0.01\\
     
     &\multicolumn{1}{|c|}{GP-VAE}&6.55$\pm$0.09 &122.33$\pm$2.05&- &
     \multicolumn{1}{|c} {6.57$\pm$0.10} &127.26$\pm$3.97 &-\\
     \cmidrule{2-8}
     &\multicolumn{1}{|c|}{CSDI}& 1.98$\pm$0.00 &12.62$\pm$0.60& -&
     \multicolumn{1}{|c} {1.79$\pm$0.00} &8.96$\pm$0.08 &\\

     &\multicolumn{1}{|c|}{RDPI}&  \textbf{1.96$\pm$0.01 }&\textbf{12.55$\pm$0.34} & \textbf{3.40$\pm$0.02} &  
     \multicolumn{1}{|c}{\textbf{1.73$\pm$0.01}}& \textbf{8.45$\pm$0.03} &\textbf{3.02$\pm$0.02}\\
     \bottomrule[1pt]
    \end{tabular}
    \caption{Imputation results on PEMS-BAY and METR-LA. Performance averaged over 5 runs.}\label{tab4}%
\end{table*}
This improvement is primarily due to our proposed conditional diffusion model, which leverages an initial model to reduce the computational cost of the denoising model. The initial model provides a quick rough imputation, whereas the conditional diffusion model efficiently estimates the residual distribution between the rough imputation and the true data.  This approach demonstrates that the RDPI framework effectively smooths the residuals between the initial imputation and the true data, thereby reducing extreme predictions and leading to more accurate and reliable results. 

\subsection{Sensitivity Analysis (RQ2)}
For spatiotemporal data, the sparsity of the data significantly affects the performance of models. We use MATR-LA to evaluate the robustness of the model under different data missing rates. The evaluation methods include BRITS, GRIN, and CSDI. We test the imputation performance of different models with data missing rates ranging from 10\% to 90\% for both patterns. For block-missing patterns, we assess the imputation performance of different models with varying missing rates over time intervals ranging [12, 48]; for point-missing patterns, we drop the observed data according to the missing rate.

\begin{figure}[h]
  \centering
  \subfigure[MAE on the METR-LA (Block)]{
    \includegraphics[width=0.47\linewidth]{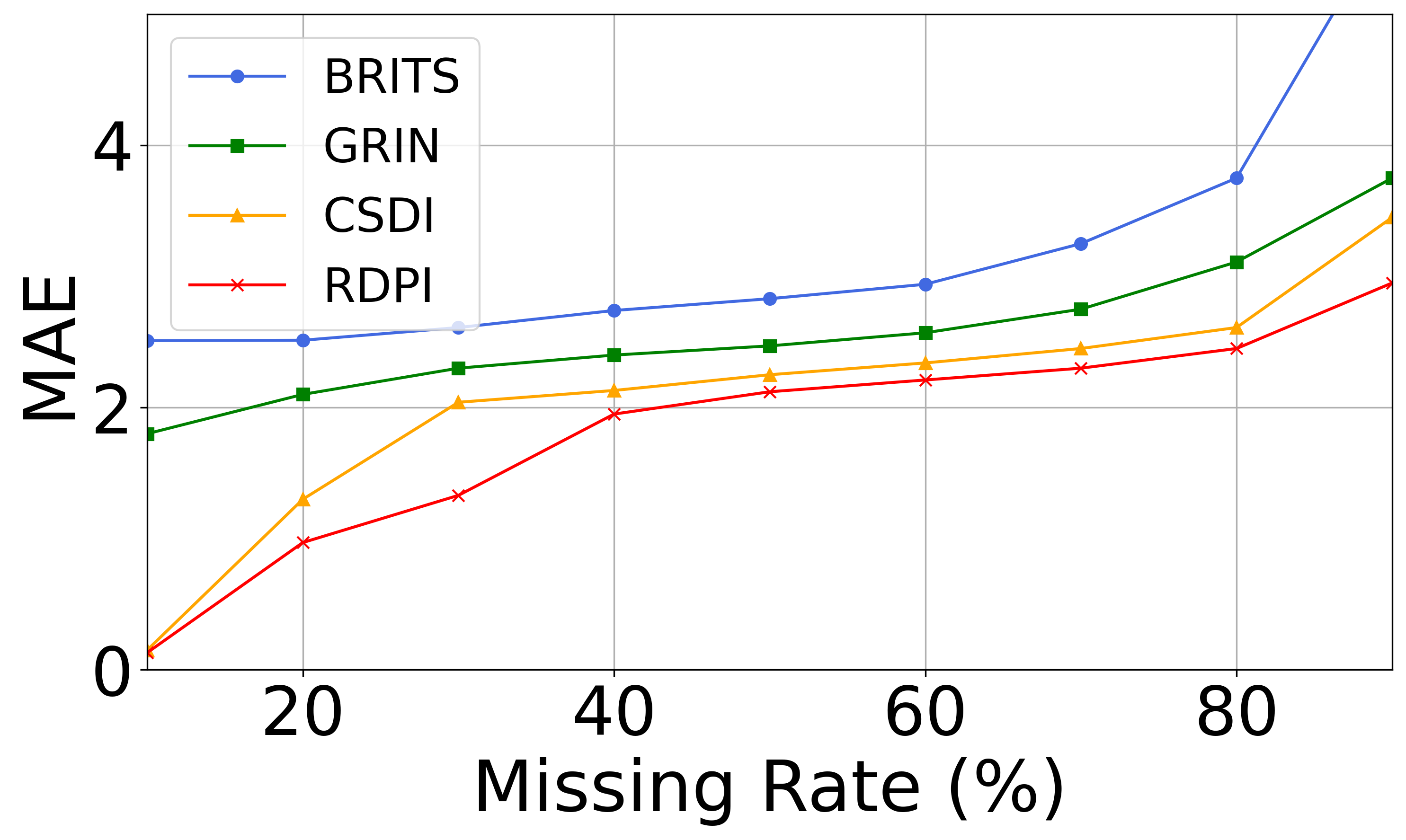}
  }
  \hfill
  \subfigure[MAE on the METR-LA (Point)]{
    \includegraphics[width=0.47\linewidth]{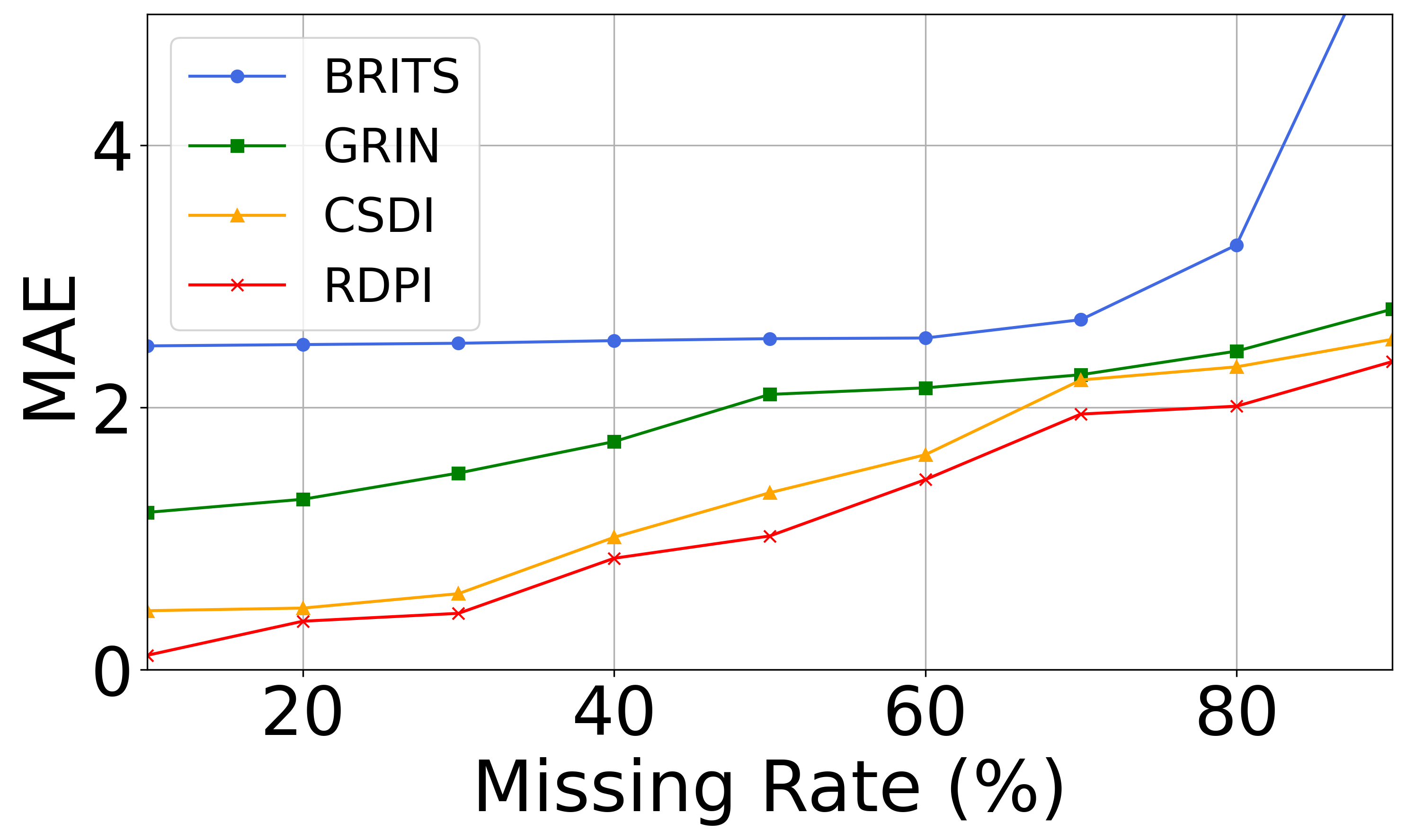}
  }
  \caption{Imputation results under different missing rates.}
  \label{miss}
\end{figure}

The MAE results for all methods are shown in Figure \ref{miss}. As illustrated in the figure, the results generated by diffusion models (RDPI and CSDI) significantly outperform those of deterministic models (BRITS and GRIN). Notably, RDPI consistently exhibits the best imputation performance even as the missing rate increases. This can be attributed to RDPI make use of observed data as a condition during the denoising and diffusion processes, which effectively captures the dependencies between observed and missing data.

\begin{table*}
\centering
    \begin{tabular}{lcccccc}
     \toprule
      \multirow{3}[3]{*}{\centering Method}& \multicolumn{3}{|c}{AQI36}& \multicolumn{3}{|c}{METR-LA}\\
     \cmidrule{2-7} 
     &\multicolumn{3}{|c}{In-sample}& \multicolumn{3}{|c}{Block-missing} \\
     \cmidrule{2-7} 
       &\multicolumn{1}{|c}{ MAE} & MSE& MRE(\%)  &\multicolumn{1}{|c}{MAE} &MSE &MRE(\%) \\
    \midrule   
    
    \multicolumn{1}{c|}{ w/o cond-forw} & 9.25$\pm$0.32 & 310.44$\pm$10.22& 12.45$\pm$4.35 &\multicolumn{1}{|c} {2.05$\pm$0.01}  & 14.73$\pm$0.03 & 4.74$\pm$0.03 \\
   
    \multicolumn{1}{c|}{w/o residual} & 8.31$\pm$4.35  &   280.57$\pm$10.03&  12.23$\pm$0.03 &\multicolumn{1}{|c} {1.97$\pm$0.01} & 13.53$\pm$0.03&4.02$\pm$0.03  \\
    \multicolumn{1}{c|}{w/o joint} & 9.20$\pm$0.01  &   263.25$\pm$0.03&  12.83$\pm$0.03 &\multicolumn{1}{|c} {2.04$\pm$0.01}  &15.01$\pm$0.02 & 4.97$\pm$0.01 \\
    \multicolumn{1}{c|}{w/o pre-train} &8.87$\pm$0.44 & 295.03$\pm$26.88 & 12.52$\pm$0.32 &\multicolumn{1}{|c} {2.12$\pm$0.01}  &16.97$\pm$0.02 & 4.21$\pm$0.01 \\
    \multicolumn{1}{c|}{predicting $x_{\theta}$} &8.29$\pm$0.32&\textbf{153.66$\pm$15.90} &11.69$\pm$0.24 & \multicolumn{1}{|c} {1.97$\pm$0.04} & 13.21$\pm$0.01&3.97$\pm$0.01 \\
     \multicolumn{1}{c|}{$- f_\theta(\mathbf{x}_0^c)$}& 9.13$\pm$0.12 &274.35$\pm$0.32& 12.64$\pm$0.32&\multicolumn{1}{|c} {2.63$\pm$0.02}&18.97$\pm$0.01   &5.24$\pm$0.02\\
   \midrule 
    \multicolumn{1}{c|}{RDPI} &\textbf{7.98$\pm$0.24}  &   238.25$\pm$13.22&  \textbf{11.67$\pm$0.35 }&\multicolumn{1}{|c} {\textbf{1.96$\pm$0.01}}&\textbf{12.55$\pm$0.34} & \textbf{3.40$\pm$0.02}  \\
    \bottomrule
    \end{tabular}

    \caption{Ablation studies on AQI36 and METR-LA.}\label{tab5}%
\end{table*}
\subsection{Abaliation Study (RQ3)}
We design ablation experiments on datasets METR-LA and AQI36 to verify the effectiveness of the conditional diffusion model, the two stage framework, and different training strategies. We compare our method with the following variants:
\begin{itemize}
    \item w/o cond-forw: the observational condition was not used in forward process.
    
    \item w/o residual:  the same denoising network was used for the true missing data.
    \item w/o joint: the initial model was frozen, and only the denoising model was trained. 
    \item w/o pre-train: the initial model was not  pretrained.
    \item predicting $x_{\theta}$: the training strategy is predicting $x_{\theta}$ rather than $\epsilon_{\theta}$.
    \item $- f_\theta(\mathbf{x}_0^c)$: the loss function calculates $\|\mathbf{x}_0^m - f_\theta(\mathbf{x}_0^c) \|$. The diffusion target is $\mathbf{x}_0^m -f_\theta(\mathbf{x}_0^c)$.
\end{itemize}  
The result of two datasets are shown in Table \ref{tab5}. As shown in the table, we have the following observations: (1) From the results of $w/o$ cond-forw, it can be observed that the imputation performance is the worst when the process is not conditioned on the observed values. This phenomenon can be explained by the fact that the lack of observed values as conditions means that the model cannot utilize the relationship between observed and missing data, thus failing to effectively capture the internal dependencies in the data. (2) The results of the residual $w/o$ indicate that the conditional diffusion model itself possesses a strong generative capacity. Even without employing a two-stage framework, relatively good imputation results can still be achieved. (3) The results of $w/o$ joint and $w/o$ pre-train show that the imputation performance of the initial model has a certain impact on the final imputation results. Specifically, $w/o$ joint shows the worst performance, which we attribute to the overfitting of residuals between the initial model and the true values on the training set due to the powerful probabilistic modeling ability of the diffusion model, leading to the poorest final imputation results. In contrast, $w/o$ pre-train demonstrates that pretraining the initial model is effective because it helps to keep the residuals between the initial model and the true values relatively stable, thereby stabilizing the training process of the diffusion model. (4) The prediction results $x_{\theta}$ indicate that the prediction of the data by the diffusion model is not the optimal choice. Our explanation is that the input to the diffusion model is the residuals. Compared to the observed values, the mean of the data is already very small, leading to a decrease in the signal-to-noise ratio and making the noise values relatively larger compared to the residuals, which makes predicting the noise more challenging. (5) The results of $- f_\theta(\mathbf{x}_0^c)$ indicate a significant decrease in imputation performance when initial model is used using the loss function $\|\mathbf{x}_0^m - f_\theta(\mathbf{x}_0^c) \|$. During the training process, it was observed that as the denoising model was optimized, the performance of the initial model's imputation deteriorated. This occurs because, with The diffusion target is $\mathbf{x}_0^m -f_\theta(\mathbf{x}_0^c)$, the loss of the initial model increases as the denoising loss decreases, leading to unstable residuals and impairing the diffusion model's ability to effectively model the residual distribution.

\subsection{Probabilistic Imputation (RQ4)}
\begin{figure}[h]
    \centering
    \includegraphics[width=0.45\textwidth]{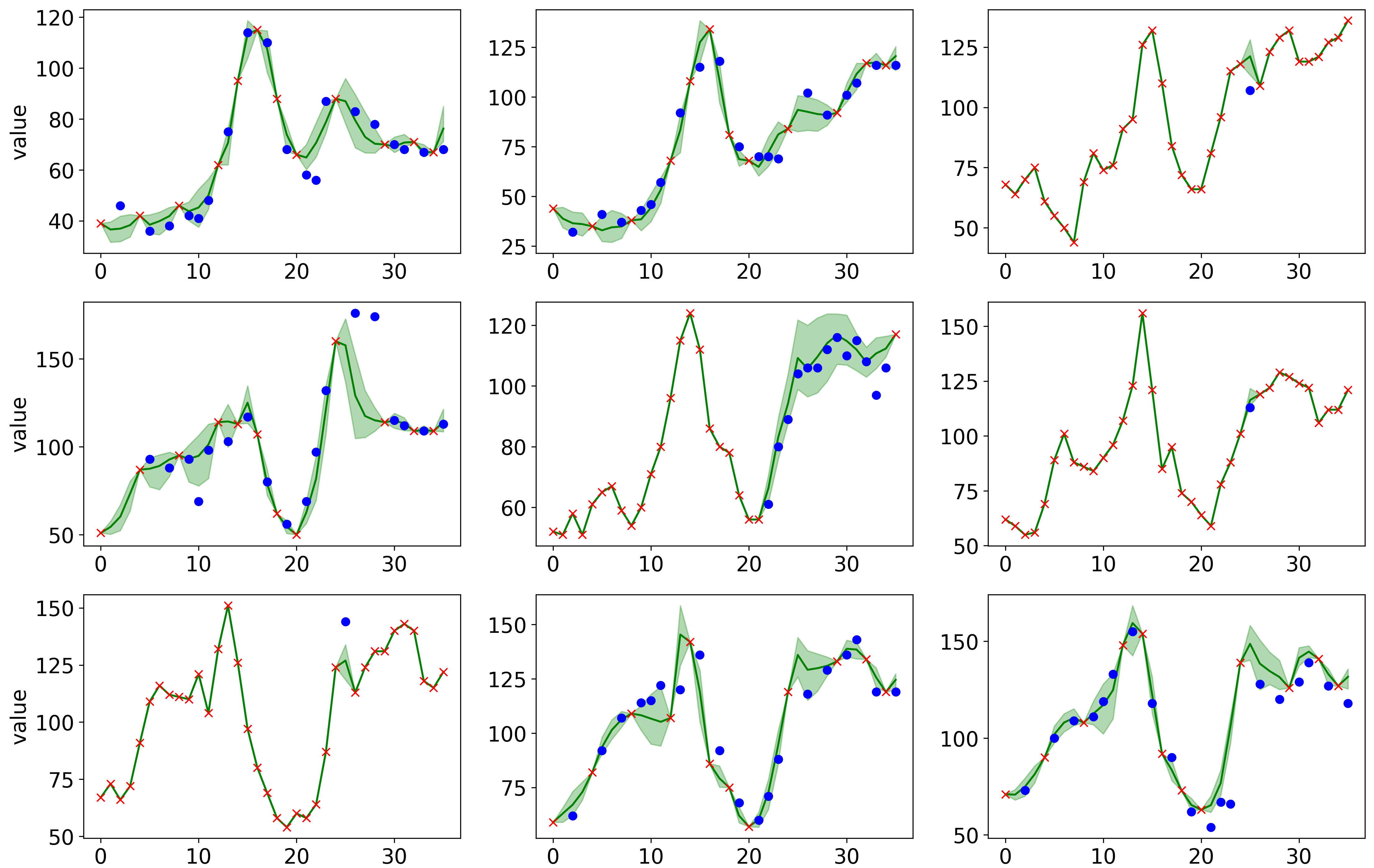}
    \caption{Imputation examples of RDPI on
the AQI36 dataset. The horizontal axis represents time, and the vertical axis represents value. }
    \label{fig3}
\end{figure}

We illustrate partial visual imputation examples of RDPI in the AQI36 data set in Figure \ref{fig3}. Red crosses denote observed values, blue circles represent missing ground truths. We conducted 50 imputations; the solid line indicates the median imputation, while the shaded green area represents the 5th and 95th percentiles. As shown, the median line generally fits the original time series well, with nearly all missing data falling within the shaded region. These results demonstrate the effectiveness of the proposed model for imputation of missing data.
\begin{figure}[h]
    \centering
    \includegraphics[width=0.45\textwidth]{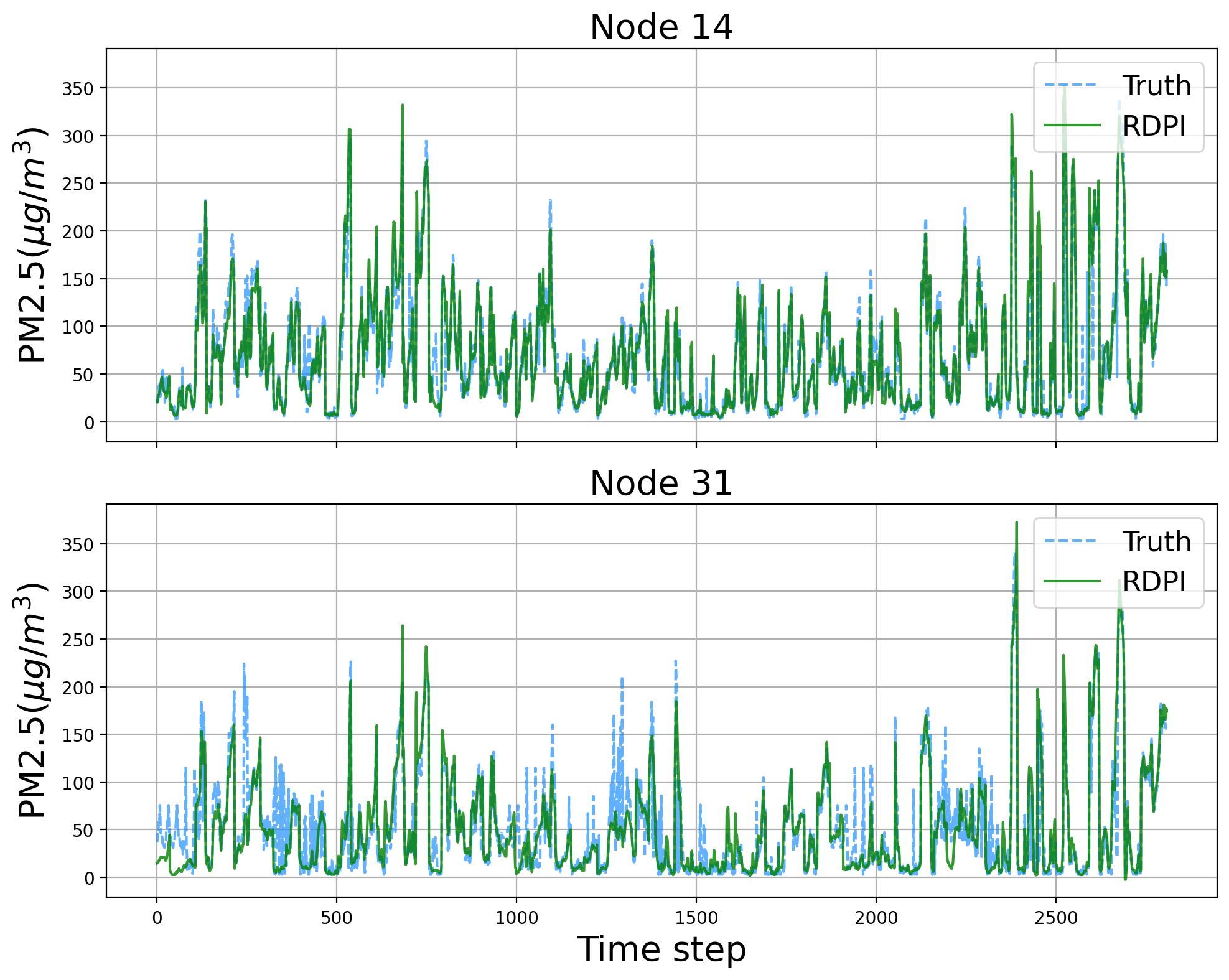}
    \caption{The imputation for unobserved sensors in AQI36. The blue dotted line represents the truth of the ground, and the green solid line represents the deterministic imputation result}
    \label{fig4}
\end{figure}

Furthermore, we investigated a scenario where a node exists but has no data and reconstructed the node's data using imputation methods. In a traffic context, such a situation might occur at locations in the network, such as entry points, where no sensor data is recorded, yet these data are crucial for traffic management. This is commonly referred to as the Kriging problem \cite{zheng2023increase}, which aims to reconstruct the time series at a specified location based on the node's geographic position and data from other observed nodes. We conducted experiments on the AQI36 dataset to address this issue. According to \cite{cini2021filling, liu2023pristi}, we selected the node with the highest connectivity (node 14) and the node with the lowest connectivity (node 31), masked the data for these nodes, and then performed imputation across the entire network. The imputation results are visualized in Figure \ref{fig4}, where the yellow lines represent the actual values and the green lines show the imputation results. The quantitative results are presented in Table \ref{tab6}.
\begin{table}
\setlength{\tabcolsep}{1mm}
    \begin{tabular}{lcccccc}
     \toprule
      & \multicolumn{3}{|c}{Node-14}& \multicolumn{3}{|c}{Node-31}\\
    
     \cmidrule{2-7} 
       &\multicolumn{1}{|c}{ MAE} & MSE& MRE(\%)  &\multicolumn{1}{|c}{MAE} &MSE &MRE(\%) \\
    \midrule   
    \multicolumn{1}{c|}{GRIN} & 13.75 & 658.33& 19.57 &\multicolumn{1}{|c} {20.55}  & 1151.35 &  39.72 \\
    \multicolumn{1}{c|}{RDPI} & \textbf{9.50}&   \textbf{316.70}& \textbf{ 13.52} &\multicolumn{1}{|c} {\textbf{15.28}} & \textbf{705.41} & \textbf{ 29.53} \\
    \bottomrule
    \end{tabular}
    \caption{Imputation result for node 14 and node 31 in AQI36.}\label{tab6}%
\end{table}
As shown in Figure \ref{fig4}, the imputed results closely match the true values. Compared to GRIN, the MAE, MSE, and MRE values for the two nodes are reduced by 31\%, 52\%, 31\% and 26\%, 39\%, 26\%. This indicates that the proposed two-stage model significantly enhances imputation performance. The conditional diffusion model acts as a residual fitting module to further optimize GRIN's imputation performance, with GRIN generating the imputation mean and the diffusion model generating the variance, thereby effectively combining the strengths of both approaches.
\subsection{Hyperparameter Analysis (RQ5)}
\begin{figure}[h]
  \centering
  \subfigure[Diffusion steps]{
    \includegraphics[width=0.3\linewidth]{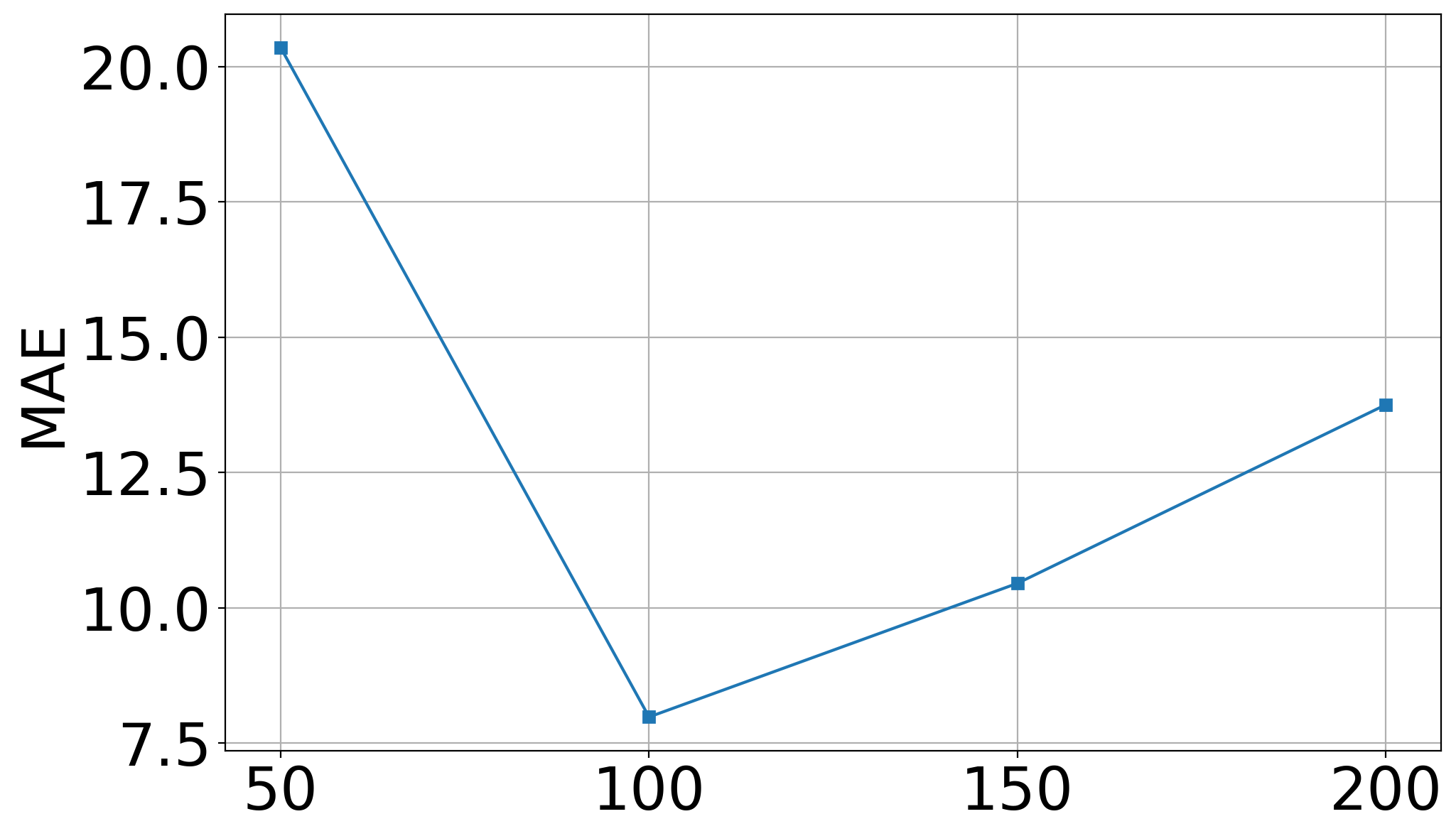}
  }
  \hfill
  \subfigure[Loss balancing parameter $\lambda$]{
    \includegraphics[width=0.3\linewidth]{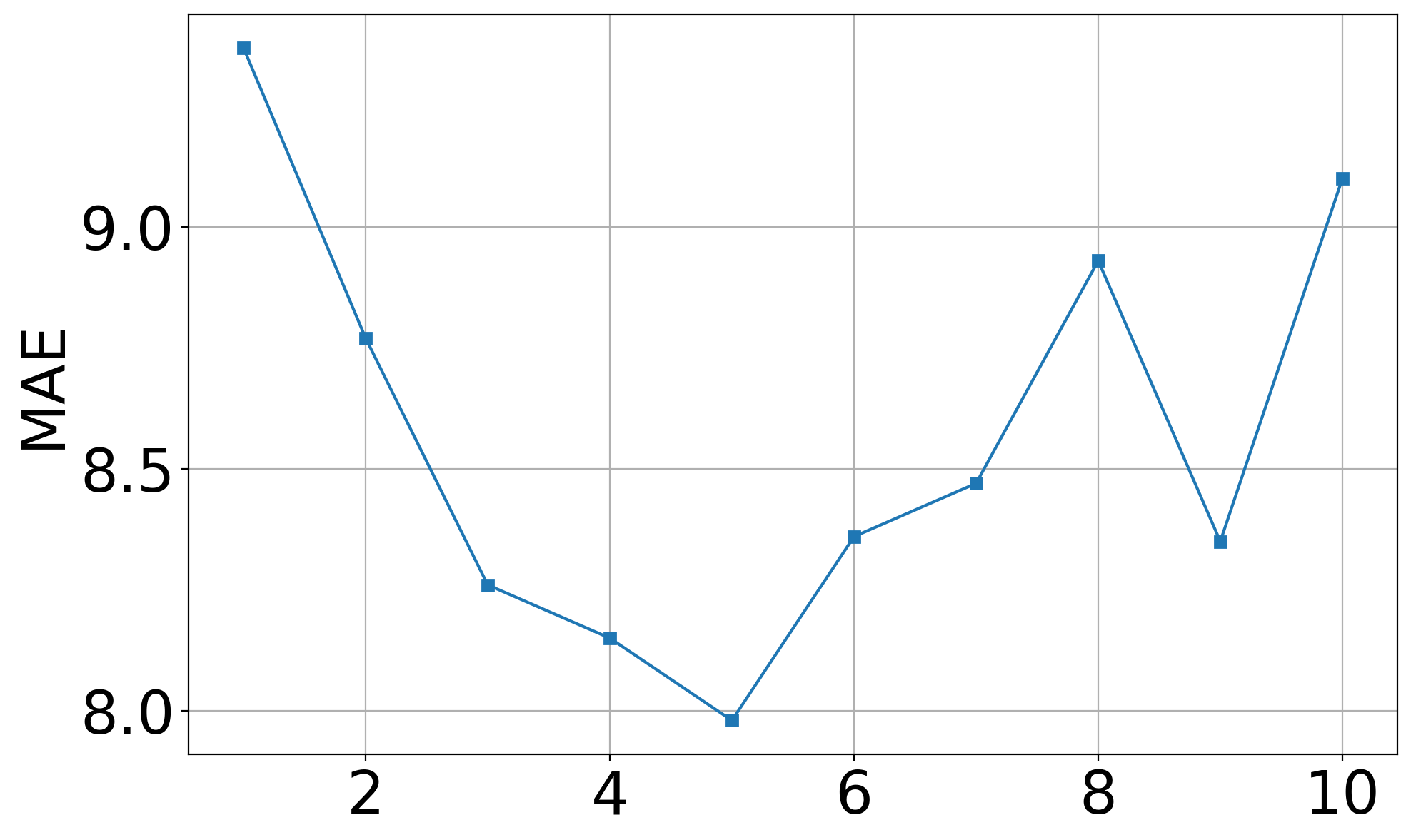}
  }
  \hfill
  \subfigure[Accelerate steps]{
    \includegraphics[width=0.3\linewidth]{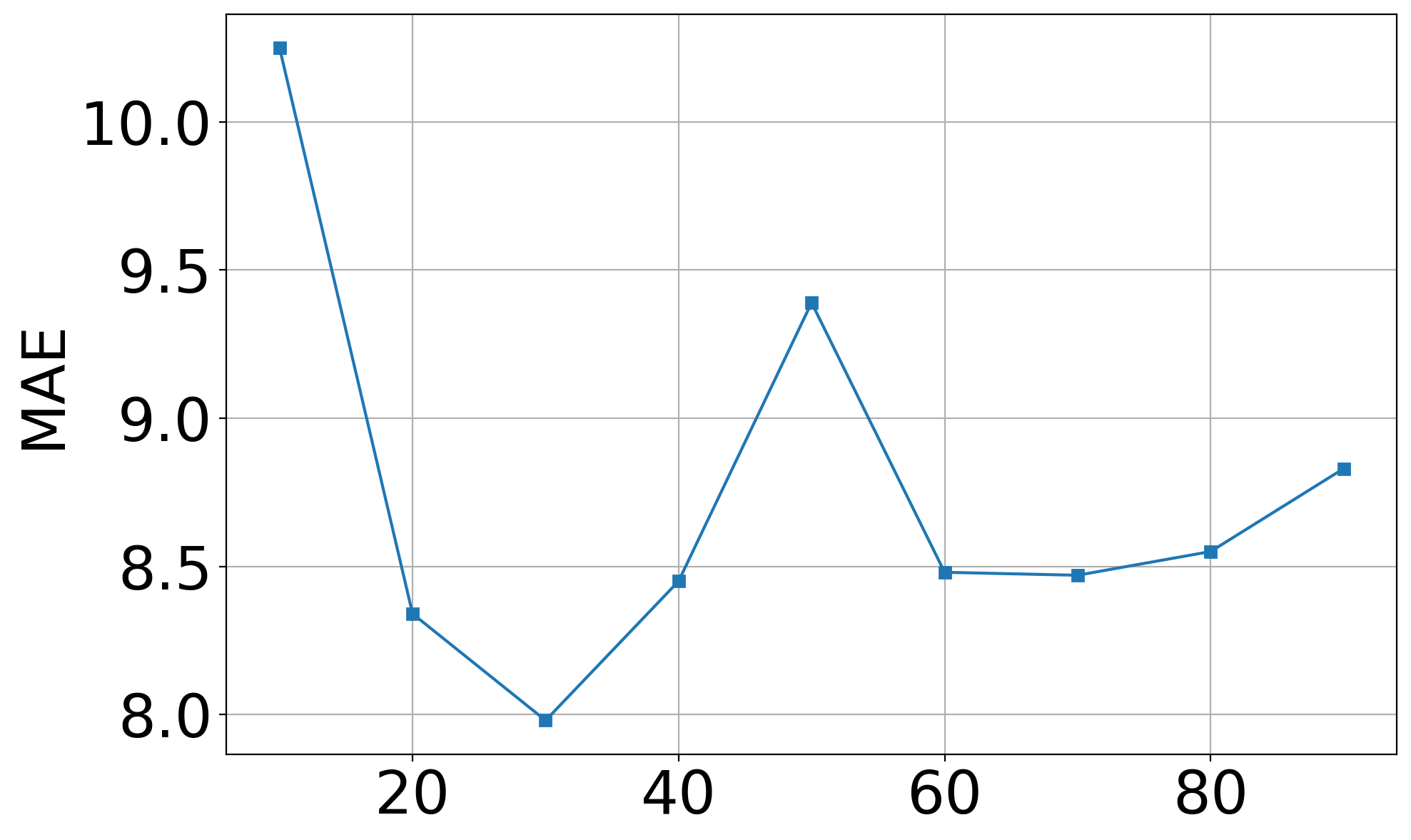}
  }
  \caption{Sensitivity analysis of key hyperparameters.}
  \label{fig7}
\end{figure}

In this section, we examine the impact of hyperparameters on the imputation performance of the model through experiments. Specifically, we tested the imputation results with different diffusion steps, different $\lambda$  and different accelerate steps in the AQI36 dataset. The result is shown in Figure \ref{fig7}. The result indicates that larger values for all parameters are not always better. Based on these findings, we empirically set different hyperparameters for different datasets, as shown in Table \ref{tab2}.

\section{Conclusion}

We presented a novel two stage diffusion probability generation method for spatiotemporal data imputation. In the initial stage, a deterministic imputation method generates coarse missing data; in the refine stage, a fully conditional diffusion model is used to reduce the gap between the coarse imputation data and the real data. Specifically, we incorporate observation conditions into the forward process, addressing the problem of estimating the probability distribution of missing data given the observed values. Based on the  defined forward and backward processes, we derived a new ELBO and proposed corresponding training and imputation methods. Extensive experiments validate the superiority and effectiveness of our method for spatiotemporal data imputation under various settings. In future work, we will focus on decoupling spatiotemporal data and exploring conditional diffusion models for graph structures and multidimensional time series to enhance the imputation and prediction performance of the existing framework in extracting spatiotemporal dependencies.
\bibliography{aaai25}

\begin{thebibliography}{37}
\providecommand{\natexlab}[1]{#1}

\bibitem[{Alcaraz and Strodthoff(2022)}]{alcaraz2022diffusion}
Alcaraz, J. M.~L.; and Strodthoff, N. 2022.
\newblock Diffusion-based time series imputation and forecasting with structured state space models.
\newblock \emph{arXiv preprint arXiv:2208.09399}.

\bibitem[{Appleby, Liu, and Liu(2020)}]{appleby2020kriging}
Appleby, G.; Liu, L.; and Liu, L.-P. 2020.
\newblock Kriging convolutional networks.
\newblock In \emph{Proceedings of the AAAI Conference on Artificial Intelligence}, volume~34, 3187--3194.

\bibitem[{Beretta and Santaniello(2016)}]{beretta2016nearest}
Beretta, L.; and Santaniello, A. 2016.
\newblock Nearest neighbor imputation algorithms: a critical evaluation.
\newblock \emph{BMC medical informatics and decision making}, 16: 197--208.

\bibitem[{Box et~al.(2015)Box, Jenkins, Reinsel, and Ljung}]{box2015time}
Box, G.~E.; Jenkins, G.~M.; Reinsel, G.~C.; and Ljung, G.~M. 2015.
\newblock \emph{Time series analysis: forecasting and control}.
\newblock John Wiley \& Sons.

\bibitem[{Cao et~al.(2018)Cao, Wang, Li, Zhou, Li, and Li}]{cao2018brits}
Cao, W.; Wang, D.; Li, J.; Zhou, H.; Li, L.; and Li, Y. 2018.
\newblock Brits: Bidirectional recurrent imputation for time series.
\newblock \emph{Advances in neural information processing systems}, 31.

\bibitem[{Che et~al.(2018)Che, Purushotham, Cho, Sontag, and Liu}]{che2018recurrent}
Che, Z.; Purushotham, S.; Cho, K.; Sontag, D.; and Liu, Y. 2018.
\newblock Recurrent neural networks for multivariate time series with missing values.
\newblock \emph{Scientific reports}, 8(1): 6085.

\bibitem[{Cini, Marisca, and Alippi(2021)}]{cini2021filling}
Cini, A.; Marisca, I.; and Alippi, C. 2021.
\newblock Filling the G\_ap\_s: Multivariate Time Series Imputation by Graph Neural Networks.
\newblock In \emph{International Conference on Learning Representations}.

\bibitem[{Du, C{\^o}t{\'e}, and Liu(2023)}]{du2023saits}
Du, W.; C{\^o}t{\'e}, D.; and Liu, Y. 2023.
\newblock Saits: Self-attention-based imputation for time series.
\newblock \emph{Expert Systems with Applications}, 219: 119619.

\bibitem[{Fortuin et~al.(2020)Fortuin, Baranchuk, R{\"a}tsch, and Mandt}]{fortuin2020gp}
Fortuin, V.; Baranchuk, D.; R{\"a}tsch, G.; and Mandt, S. 2020.
\newblock Gp-vae: Deep probabilistic time series imputation.
\newblock In \emph{International conference on artificial intelligence and statistics}, 1651--1661. PMLR.

\bibitem[{Han, Zheng, and Zhou(2022)}]{han2022card}
Han, X.; Zheng, H.; and Zhou, M. 2022.
\newblock Card: Classification and regression diffusion models.
\newblock \emph{Advances in Neural Information Processing Systems}, 35: 18100--18115.

\bibitem[{Hastie et~al.(2009)Hastie, Tibshirani, Friedman, and Friedman}]{hastie2009elements}
Hastie, T.; Tibshirani, R.; Friedman, J.~H.; and Friedman, J.~H. 2009.
\newblock \emph{The elements of statistical learning: data mining, inference, and prediction}, volume~2.
\newblock Springer.

\bibitem[{Ho, Jain, and Abbeel(2020)}]{ho2020denoising}
Ho, J.; Jain, A.; and Abbeel, P. 2020.
\newblock Denoising diffusion probabilistic models.
\newblock \emph{Advances in neural information processing systems}, 33: 6840--6851.

\bibitem[{Hu et~al.(2023)Hu, Liang, Fan, Chen, Zheng, and Zimmermann}]{hu2023graph}
Hu, J.; Liang, Y.; Fan, Z.; Chen, H.; Zheng, Y.; and Zimmermann, R. 2023.
\newblock Graph Neural Processes for Spatio-Temporal Extrapolation.
\newblock In \emph{Proceedings of the 29th ACM SIGKDD Conference on Knowledge Discovery and Data Mining}, 752--763.

\bibitem[{Kim et~al.(2023)Kim, Kim, Yun, Lee, Lee, and Lee}]{kim2023probabilistic}
Kim, S.; Kim, H.; Yun, E.; Lee, H.; Lee, J.; and Lee, J. 2023.
\newblock Probabilistic imputation for time-series classification with missing data.
\newblock In \emph{International Conference on Machine Learning}, 16654--16667. PMLR.

\bibitem[{Li et~al.(2024)Li, Chen, Hu, Chen, baolin sun, and Zhou}]{li2024transformermodulated}
Li, Y.; Chen, W.; Hu, X.; Chen, B.; baolin sun; and Zhou, M. 2024.
\newblock Transformer-Modulated Diffusion Models for Probabilistic Multivariate Time Series Forecasting.
\newblock In \emph{The Twelfth International Conference on Learning Representations}.

\bibitem[{Li et~al.(2017)Li, Yu, Shahabi, and Liu}]{li2017diffusion}
Li, Y.; Yu, R.; Shahabi, C.; and Liu, Y. 2017.
\newblock Diffusion convolutional recurrent neural network: Data-driven traffic forecasting.
\newblock \emph{arXiv preprint arXiv:1707.01926}.

\bibitem[{Liu et~al.(2023)Liu, Huang, Feng, Sun, Du, and Fu}]{liu2023pristi}
Liu, M.; Huang, H.; Feng, H.; Sun, L.; Du, B.; and Fu, Y. 2023.
\newblock Pristi: A conditional diffusion framework for spatiotemporal imputation.
\newblock In \emph{2023 IEEE 39th International Conference on Data Engineering (ICDE)}, 1927--1939. IEEE.

\bibitem[{Liu et~al.(2019)Liu, Yu, Zheng, Zhan, and Yue}]{liu2019naomi}
Liu, Y.; Yu, R.; Zheng, S.; Zhan, E.; and Yue, Y. 2019.
\newblock Naomi: Non-autoregressive multiresolution sequence imputation.
\newblock \emph{Advances in neural information processing systems}, 32.

\bibitem[{Luo et~al.(2018)Luo, Cai, Zhang, Xu et~al.}]{luo2018multivariate}
Luo, Y.; Cai, X.; Zhang, Y.; Xu, J.; et~al. 2018.
\newblock Multivariate time series imputation with generative adversarial networks.
\newblock \emph{Advances in neural information processing systems}, 31.

\bibitem[{Marisca, Cini, and Alippi(2022)}]{marisca2022learning}
Marisca, I.; Cini, A.; and Alippi, C. 2022.
\newblock Learning to reconstruct missing data from spatiotemporal graphs with sparse observations.
\newblock \emph{Advances in Neural Information Processing Systems}, 35: 32069--32082.

\bibitem[{Mattei and Frellsen(2019)}]{mattei2019miwae}
Mattei, P.-A.; and Frellsen, J. 2019.
\newblock MIWAE: Deep generative modelling and imputation of incomplete data sets.
\newblock In \emph{International conference on machine learning}, 4413--4423. PMLR.

\bibitem[{Miao et~al.(2021)Miao, Wu, Wang, Gao, Mao, and Yin}]{miao2021generative}
Miao, X.; Wu, Y.; Wang, J.; Gao, Y.; Mao, X.; and Yin, J. 2021.
\newblock Generative semi-supervised learning for multivariate time series imputation.
\newblock In \emph{Proceedings of the AAAI conference on artificial intelligence}, volume~35, 8983--8991.

\bibitem[{Mulyadi, Jun, and Suk(2021)}]{mulyadi2021uncertainty}
Mulyadi, A.~W.; Jun, E.; and Suk, H.-I. 2021.
\newblock Uncertainty-aware variational-recurrent imputation network for clinical time series.
\newblock \emph{IEEE Transactions on Cybernetics}, 52(9): 9684--9694.

\bibitem[{Shan, Li, and Oliva(2023)}]{shan2023nrtsi}
Shan, S.; Li, Y.; and Oliva, J.~B. 2023.
\newblock Nrtsi: Non-recurrent time series imputation.
\newblock In \emph{ICASSP 2023-2023 IEEE International Conference on Acoustics, Speech and Signal Processing (ICASSP)}, 1--5. IEEE.

\bibitem[{Song, Meng, and Ermon(2021)}]{songdenoising}
Song, J.; Meng, C.; and Ermon, S. 2021.
\newblock Denoising Diffusion Implicit Models.
\newblock In \emph{International Conference on Learning Representations}.

\bibitem[{Tashiro et~al.(2021)Tashiro, Song, Song, and Ermon}]{tashiro2021csdi}
Tashiro, Y.; Song, J.; Song, Y.; and Ermon, S. 2021.
\newblock Csdi: Conditional score-based diffusion models for probabilistic time series imputation.
\newblock \emph{Advances in Neural Information Processing Systems}, 34: 24804--24816.

\bibitem[{Vaswani et~al.(2017)Vaswani, Shazeer, Parmar, Uszkoreit, Jones, Gomez, Kaiser, and Polosukhin}]{vaswani2017attention}
Vaswani, A.; Shazeer, N.; Parmar, N.; Uszkoreit, J.; Jones, L.; Gomez, A.~N.; Kaiser, {\L}.; and Polosukhin, I. 2017.
\newblock Attention is all you need.
\newblock \emph{Advances in neural information processing systems}, 30.

\bibitem[{Wang et~al.(2023)Wang, Zhang, Wang, Zhang, Wang, Zhou, and Wang}]{wang2023observed}
Wang, X.; Zhang, H.; Wang, P.; Zhang, Y.; Wang, B.; Zhou, Z.; and Wang, Y. 2023.
\newblock An observed value consistent diffusion model for imputing missing values in multivariate time series.
\newblock In \emph{Proceedings of the 29th ACM SIGKDD Conference on Knowledge Discovery and Data Mining}, 2409--2418.

\bibitem[{Whang et~al.(2022)Whang, Delbracio, Talebi, Saharia, Dimakis, and Milanfar}]{whang2022deblurring}
Whang, J.; Delbracio, M.; Talebi, H.; Saharia, C.; Dimakis, A.~G.; and Milanfar, P. 2022.
\newblock Deblurring via stochastic refinement.
\newblock In \emph{Proceedings of the IEEE/CVF Conference on Computer Vision and Pattern Recognition}, 16293--16303.

\bibitem[{White, Royston, and Wood(2011)}]{white2011multiple}
White, I.~R.; Royston, P.; and Wood, A.~M. 2011.
\newblock Multiple imputation using chained equations: issues and guidance for practice.
\newblock \emph{Statistics in medicine}, 30(4): 377--399.

\bibitem[{Wu et~al.(2021)Wu, Zhuang, Labbe, and Sun}]{wu2021inductive}
Wu, Y.; Zhuang, D.; Labbe, A.; and Sun, L. 2021.
\newblock Inductive graph neural networks for spatiotemporal kriging.
\newblock In \emph{Proceedings of the AAAI Conference on Artificial Intelligence}, volume~35, 4478--4485.

\bibitem[{Yi et~al.(2016)Yi, Zheng, Zhang, and Li}]{yi2016st}
Yi, X.; Zheng, Y.; Zhang, J.; and Li, T. 2016.
\newblock ST-MVL: Filling missing values in geo-sensory time series data.
\newblock In \emph{Proceedings of the 25th International Joint Conference on Artificial Intelligence}.

\bibitem[{Yoon, Jordon, and Schaar(2018)}]{yoon2018gain}
Yoon, J.; Jordon, J.; and Schaar, M. 2018.
\newblock Gain: Missing data imputation using generative adversarial nets.
\newblock In \emph{International conference on machine learning}, 5689--5698. PMLR.

\bibitem[{Yoon, Zame, and van~der Schaar(2018)}]{yoon2018estimating}
Yoon, J.; Zame, W.~R.; and van~der Schaar, M. 2018.
\newblock Estimating missing data in temporal data streams using multi-directional recurrent neural networks.
\newblock \emph{IEEE Transactions on Biomedical Engineering}, 66(5): 1477--1490.

\bibitem[{You et~al.(2020)You, Ma, Ding, Kochenderfer, and Leskovec}]{you2020handling}
You, J.; Ma, X.; Ding, Y.; Kochenderfer, M.~J.; and Leskovec, J. 2020.
\newblock Handling missing data with graph representation learning.
\newblock \emph{Advances in Neural Information Processing Systems}, 33: 19075--19087.

\bibitem[{Zheng et~al.(2023)Zheng, Fan, Wang, Qi, Chen, and Chen}]{zheng2023increase}
Zheng, C.; Fan, X.; Wang, C.; Qi, J.; Chen, C.; and Chen, L. 2023.
\newblock Increase: Inductive graph representation learning for spatio-temporal kriging.
\newblock In \emph{Proceedings of the ACM Web Conference 2023}, 673--683.

\bibitem[{Zivot and Wang(2006)}]{zivot2006vector}
Zivot, E.; and Wang, J. 2006.
\newblock Vector autoregressive models for multivariate time series.
\newblock \emph{Modeling financial time series with S-PLUS{\textregistered}}, 385--429.

\end{thebibliography}


\appendix
\section{Appendix}
\subsection{Derivation of ELBO for Forward Process Posteriors}
\label{sec:a1}
In this section, we derive the expressions for the ELBO of the conditional distribution $ p(\mathbf{z}_0^m|\mathbf{z}_0^c)$ and the posterior probability of the forward process.
\begin{equation}
\label{qeq1}
\begin{aligned}
    &\log p(\mathbf{z}_0^m|\mathbf{z}_0^c)  = \log \int p(\mathbf{z}_{0:T}^m|\mathbf{z}_0^c) d\mathbf{z}_{1:T}^m \\
                        & = \log \int \frac{p(\mathbf{z}_{0:T}^m|\mathbf{z}_0^c)} {q(\mathbf{z}_{1:T}^m|\mathbf{z}_0^m,\mathbf{z}_0^c)}q(\mathbf{z}_{1:T}^m|\mathbf{z}_0^m,\mathbf{z}_0^c)d\mathbf{z}_{1:T}^m \\
                        & = \log \mathbb{E}_{q(\mathbf{z}_{1:T}^m|\mathbf{z}_0^m,\mathbf{z}_0^c)}[\frac{p(\mathbf{z}_{0:T}^m|\mathbf{z}_0^c)} {q(\mathbf{z}_{1:T}^m|\mathbf{z}_0^m,\mathbf{z}_0^c)}] \\
                        & \geq \mathbb{E}_{q(\mathbf{z}_{1:T}^m|\mathbf{z}_0^m,\mathbf{z}_0^c)}[\log \frac{p(\mathbf{z}_{0:T}^m|\mathbf{z}_0^c)} {q(\mathbf{z}_{1:T}^m|\mathbf{z}_0^m,\mathbf{z}_0^c)}]\\
                        & = \mathbb{E}_{q(\mathbf{z}_{1:T}^m|\mathbf{z}_0^m,\mathbf{z}_0^c)}[\log\frac{p(\mathbf{z}_T^m|\mathbf{z}_0^c)\prod_{t=1}^T p(\mathbf{z}_{t-1}^m|\mathbf{z}_t^m, \mathbf{z}_0^c)}{\prod_{t=1}^T q(\mathbf{z}_t^m|\mathbf{z}_{t-1}^m, \mathbf{z}_0^c)}] \\
                        & = \mathbb{E}_{q(\mathbf{z}_{1:T}^m|\mathbf{z}_0^m,\mathbf{z}_0^c)}[\log p(\mathbf{z}_T^m|\mathbf{z}_0^c)+ \log \frac{p(\mathbf{z}_0^m|\mathbf{z}_1^m,\mathbf{z}_0^c)}{q(\mathbf{z}_1^m|\mathbf{z}_0^m,\mathbf{z}_0^c)}  \\
                        &+\sum_{t=2}^T \log\frac{ p(\mathbf{z}_{t-1}^m|\mathbf{z}_t^m, \mathbf{z}_0^c)}{ q(\mathbf{z}_t^m|\mathbf{z}_{t-1}^m, \mathbf{z}_0^c)}] \\
                        & = \mathbb{E}_{q(\mathbf{z}_{1:T}^m|\mathbf{z}_0^m,\mathbf{z}_0^c)}[\log p(\mathbf{z}_T^m|\mathbf{z}_0^c)+ \log \frac{p(\mathbf{z}_0^m|\mathbf{z}_1^m,\mathbf{z}_0^c)}{q(\mathbf{z}_1^m|\mathbf{z}_0^m,\mathbf{z}_0^c)} \\
                        &+ \sum_{t=2}^T \log\frac{ p(\mathbf{z}_{t-1}^m|\mathbf{z}_t^m, \mathbf{z}_0^c)}{ q(\mathbf{z}_{t-1}^m|\mathbf{z}_t^m,\mathbf{z}_0^m, \mathbf{z}_0^c)} \cdot \frac{q(\mathbf{z}_{t-1}^m|\mathbf{z}_0^m, \mathbf{z}_0^c)}{q(\mathbf{z}_t^m|\mathbf{z}_0^m, \mathbf{z}_0^c)}] \\
                        & = \mathbb{E}_{q(\mathbf{z}_{1:T}^m|\mathbf{z}_0^m,\mathbf{z}_0^c)}[\log p(\mathbf{z}_T^m|\mathbf{z}_0^c)+ \log \frac{p(\mathbf{z}_0^m|\mathbf{z}_1^m,\mathbf{z}_0^c)}{q(\mathbf{z}_1^m|\mathbf{z}_0^m,\mathbf{z}_0^c)} \\
                        &+ \sum_{t=2}^T \log\frac{ p(\mathbf{z}_{t-1}^m|\mathbf{z}_t^m, \mathbf{z}_0^c)}{ q(\mathbf{z}_{t-1}^m|\mathbf{z}_t^m,\mathbf{z}_0^m, \mathbf{z}_0^c)} + \log \frac{q(\mathbf{z}_1^m|\mathbf{z}_0^m,\mathbf{z}_0^c)}{q(\mathbf{z}_T^m|\mathbf{z}_0^m,\mathbf{z}_0^c)}] \\
                        & = \mathbb{E}_{q(\mathbf{z}_{1:T}^m|\mathbf{z}_0^m,\mathbf{z}_0^c)}[\log \frac{p(\mathbf{z}_T^m|\mathbf{z}_0^c)}{q(\mathbf{z}_T^m|\mathbf{z}_0^m,\mathbf{z}_0^c)}+ \log p(\mathbf{z}_0^m|\mathbf{z}_1^m,\mathbf{z}_0^c)  \\
                        &+ \sum_{t=2}^T \log\frac{ p(\mathbf{z}_{t-1}^m|\mathbf{z}_t^m, \mathbf{z}_0^c)}{ q(\mathbf{z}_{t-1}^m|\mathbf{z}_t^m,\mathbf{z}_0^m, \mathbf{z}_0^c)} ] \\
                        & = \mathbb{E}_{q(\mathbf{z}_{1:T}^m|\mathbf{z}_0^m,\mathbf{z}_0^c)}[D_{KL}(q(\mathbf{z}_T^m|\mathbf{z}_0^m,\mathbf{z}_0^c)||p(\mathbf{z}_T^m|\mathbf{z}_0^c)) \\
                        &+\sum_{i=2}^TD_{KL}(q(\mathbf{z}_{t-1}^m|\mathbf{z}_t^m,\mathbf{z}_0^m, \mathbf{z}_0^c)||p(\mathbf{z}_{t-1}^m|\mathbf{z}_t^m, \mathbf{z}_0^c)) 
                          \\ &+ \log p(\mathbf{z}_0^m|\mathbf{z}_1^m,\mathbf{z}_0^c)]
\end{aligned}
\end{equation}
\subsection{Probability Distribution of Reverse Process}
\label{sec:a2}
Using Bayes' theorem and the standard Gaussian distribution probability density function, we can derive the expression $q(\mathbf{z}_{t-1}^m|\mathbf{z}_t^m,\mathbf{z}_0^m, \mathbf{z}_0^c)$ 
 from Eq.\eqref{eq6}:
\begin{equation}
\begin{aligned}
        &q(\mathbf{z}_{t-1}^m|\mathbf{z}_t^m,\mathbf{z}_0^m, \mathbf{z}_0^c) 
        = q(\mathbf{z}_t^m|\mathbf{z}_{t-1}^m, \mathbf{z}_0^m,\mathbf{z}_0^c)\frac{q(\mathbf{z}_{t-1}^m|\mathbf{z}_0^m,\mathbf{z}_0^c)}{q(\mathbf{z}_t^m|\mathbf{z}_0^m,\mathbf{z}_0^c)} \\
        &= \mathcal{N}(\mathbf{z}_t^m;\sqrt{\overline{\alpha}_t}(\mathbf{z}_{t-1}^m+\mathbf{z}_0^c),\beta_t\mathbf{I}) 
         \cdot \\
         &\frac{\mathcal{N}(\mathbf{z}_{t-1}^m;\sqrt{{\alpha_{t-1}}}(\mathbf{z}_0^m+\mathbf{z}_0^c),(1-\alpha_{t-1})\mathbf{I})}{\mathcal{N}(\mathbf{z}_t^m;\sqrt{\alpha_t}(\mathbf{z}_0^m+\mathbf{z}_0^c),(1-\alpha_t)\mathbf{I})} \\
        &\propto exp( -\frac{1}{2}(\frac{(\mathbf{z}_t^m-\sqrt{\overline{\alpha}_t}(\mathbf{z}_{t-1}^m+\mathbf{z}_0^c))^2}{\beta_t} + \\
        &
        \frac{(\mathbf{z}_{t-1}^m-\sqrt{\alpha_{t-1}}(\mathbf{z}_0^m+\mathbf{z}_0^c))^2}{1-\alpha_{t-1}})) \\
\end{aligned}
\end{equation}  

\begin{equation}
\begin{aligned}
        &= exp( -\frac{1}{2}(\frac{{\mathbf{z}_t^m}^2-2\mathbf{z}_t^m\sqrt{\overline{\alpha}_t}(\mathbf{z}_{t-1}^m+\mathbf{z}_0^c) +(\sqrt{\overline{\alpha}_t}(\mathbf{z}_{t-1}^m+\mathbf{z}_0^c))^2}{\beta_t} \\
        &+ \frac{{\mathbf{z}_{t-1}^m}^2-2\mathbf{z}_{t-1}^m\sqrt{{\alpha_{t-1}}}(\mathbf{z}_0^m+\mathbf{z}_0^c) +(\sqrt{{\alpha_{t-1}}}(\mathbf{z}_0^m+\mathbf{z}_0^c))^2}{1-\alpha_{t-1}})) \\
\end{aligned}
\end{equation}
\begin{equation}
\begin{aligned}
        &= exp( -\frac{1}{2}( (\frac{\overline{\alpha}_t}{\beta_t} +\frac{1}{1-\alpha_{t-1}}){\mathbf{z}_{t-1}^m}^2 -(\frac{2\sqrt{\overline{\alpha}_t}\mathbf{z}_t^m - 2\overline{\alpha}_t\mathbf{z}_0^c}{\beta_t} + \\        &\frac{2\sqrt{{\alpha_{t-1}}}\mathbf{z}_0^m+2\sqrt{{\alpha_{t-1}}}\mathbf{z}_0^c}{1-\alpha_{t-1}}){\mathbf{z}_{t-1}^m})) + C(\mathbf{z}_t^m,\mathbf{z}_0^m,\mathbf{z}_0^c ) \\
\end{aligned}
\end{equation}

\subsection{Mean of Reverse Process}
\label{sec:a3}
Following the standard Gaussian density function, the mean in Eq.\eqref{eq7} can be parameterized as follows 
\begin{equation}
\begin{aligned}
    &\widetilde{\mathbf{\mu}}_t
    (\mathbf{z}_{t}^m,\mathbf{z}_0^c,t)\\
    &= (\frac{\sqrt{\overline{\alpha}_t}\mathbf{z}_t^m - \overline{\alpha}_t\mathbf{z}_0^c}{\beta_t} + \frac{\sqrt{{\alpha_{t-1}}}(\mathbf{z}_0^m+\mathbf{z}_0^c)}{1-\alpha_{t-1}})/(\frac{\overline{\alpha}_t}{\beta_t} +\frac{1}{1-\alpha_{t-1}}) \\
    &= (\frac{\sqrt{\overline{\alpha}_t}\mathbf{z}_t^m - \overline{\alpha}_t\mathbf{z}_0^c}{\beta_t} + \frac{\sqrt{{\alpha_{t-1}}}(\mathbf{z}_0^m+\mathbf{z}_0^c)}{1-\alpha_{t-1}})\frac{1-\alpha_{t-1}}{1-\alpha_t}\beta_t\\
\end{aligned}
\end{equation}
\begin{equation}
\begin{aligned}    
    &= \frac{(\sqrt{\overline{\alpha}_t}\mathbf{z}_t^m - \overline{\alpha}_t\mathbf{z}_0^c)(1-\alpha_{t-1})}{1-\alpha_t} + \frac{(\sqrt{{\alpha_{t-1}}}(\mathbf{z}_0^m+\mathbf{z}_0^c))\beta_t}{1-\alpha_t} \\
    &= \frac{\sqrt{\overline{\alpha}_t}(1-\alpha_{t-1})}{1-\alpha_t}\mathbf{z}_t^m  
    +\frac{\sqrt{\alpha_{t-1}}\beta_t}{1-\alpha_t}\mathbf{z}_0^m +\\
    &\frac{\sqrt{{\alpha_{t-1}}}\beta_t- \overline{\alpha}_t(1-\alpha_{t-1})}{1-\alpha_t}\mathbf{z}_0^c 
\end{aligned}
\end{equation}

\subsection{Representation of Mean }
\label{sec:a4}
Using represent $\mathbf{z}_0^m = (\mathbf{z}_t^m - \sqrt{\alpha_t}\mathbf{z}_0^c - \sqrt{(1-\alpha_t)} \mathbf{\epsilon}_t)/\sqrt{\alpha_t}$, we can get new expression of $\widetilde{\mathbf{\mu}}_t(\mathbf{z}_{t}^m,\mathbf{z}_0^c,t)$ in Eq.\eqref{eq9}
\begin{equation}
\begin{aligned}
    &\widetilde{\mathbf{\mu}}_t(\mathbf{z}_{t}^m,\mathbf{z}_0^c,t)\\
    &= \frac{\sqrt{\overline{\alpha}_t}(1-\alpha_{t-1})}{1-\alpha_t}\mathbf{z}_t^m  \\
    &
    +\frac{\sqrt{\alpha_{t-1}}\beta_t}{1-\alpha_t}\frac{\mathbf{z}_t^m - \sqrt{\alpha_t}\mathbf{z}_0^c - \sqrt{(1-\alpha_t)} \mathbf{\epsilon}_t}{\sqrt{\alpha_t}} \\
    &+ \frac{\sqrt{{\alpha_{t-1}}}\beta_t- \overline{\alpha}_t(1-\alpha_{t-1})}{1-\alpha_t}\mathbf{z}_0^c  \\
    &= (\frac{\sqrt{\overline{\alpha}_t}(1-\alpha_{t-1})}{1-\alpha_t} + \frac{1-\overline{\alpha}_t}{(1-\alpha_t)\sqrt{\overline{\alpha}_t}})\mathbf{z}_t^m - \frac{1-\overline{\alpha}_t}{\sqrt{(1-\alpha_t) \overline{\alpha}_t}}\mathbf{\epsilon}_t\\
    &-\frac{\beta_t\sqrt{\alpha_{t-1}}}{1-\alpha_t}\mathbf{z}_0^c+\frac{\sqrt{{\alpha_{t-1}}}\beta_t- \overline{\alpha}_t(1-\alpha_{t-1})}{1-\alpha_t}\mathbf{z}_0^c\\
    &=\frac{1}{\sqrt{\overline{\alpha}_t}}\mathbf{z}_t^m -\frac{1-\overline{\alpha}_t}{\sqrt{(1-\alpha_t) \overline{\alpha}_t}}\mathbf{\epsilon}_t -\frac{\overline{\alpha}_t(1-\alpha_{t-1})}{1-\alpha_t}\mathbf{z}_0^c\\
    &= \frac{1}{\sqrt{\overline{\alpha}_t}}(\mathbf{z}_t^m  -\frac{\overline{\alpha}_t\sqrt{\overline{\alpha}_t}(1-\alpha_{t-1})}{1-\alpha_t}\mathbf{z}_0^c-\frac{1-\overline{\alpha}_t}{\sqrt{(1-\alpha_t) }}\mathbf{\epsilon}_t)\\
\end{aligned}
\end{equation}

\subsection{Diffusion Loss Function}
\label{sec:a5}
\begin{equation}
\begin{aligned}
    L_t &= \mathbb{E}_{\mathbf{z}_0^m, \epsilon}[\frac{1}{2\|\mathbf{\Sigma}_{\theta}(\mathbf{z}_{t}^m,\mathbf{z}_0^c,t) \|_2^2}\|\widetilde{\mathbf{\mu}}_t- \mathbf{\mu}_{\theta}\|]\\
    &= \mathbb{E}_{\mathbf{z}_0^m, \epsilon}[\frac{1}{2\|\mathbf{\Sigma}_{\theta} \|_2^2}\|\frac{1}{\sqrt{\overline{\alpha}_t}}(\mathbf{z}_t^m + (\sqrt{\overline{\alpha}_t}-1)\mathbf{z}_0^c \\
    &-\frac{1-\overline{\alpha}_t}{\sqrt{(1-\alpha_t) }}\mathbf{\epsilon}_t)- \frac{1}{\sqrt{\overline{\alpha}_t}}(\mathbf{z}_t^m + (\sqrt{\overline{\alpha}_t}-1)\mathbf{z}_0^c \\
    &-\frac{1-\overline{\alpha}_t}{\sqrt{(1-\alpha_t) }}\mathbf{\epsilon}_{\theta}(\mathbf{z}_t^m, \mathbf{z}_0^c, \mathbf{z}_0^c, t))\|]\\
    &= \mathbb{E}_{\mathbf{z}_0^m, \epsilon}[\frac{(1-\overline{\alpha}_t)^2}{2(1-\alpha_t)\|\mathbf{\Sigma}_{\theta} \|_2^2}\|\mathbf{\epsilon}_t- \mathbf{\epsilon}_{\theta}(\mathbf{z}_t^m, \mathbf{z}_0^c,\mathbf{z}_0^c, t)\|]\\
\end{aligned}
\end{equation}

\subsection{Accelerated Cnditional Diffusion Model}
\label{subsec:a6}
According to Eq.\eqref{eq5}, one can generate $\mathbf{z}_{t-1}^m$ and $\mathbf{z}_t^m$ from $\mathbf{z}_0^m$ and $\mathbf{z}_0^c$ via:
\begin{equation}
\begin{aligned}
    &\mathbf{z}_{t-1}^m = \sqrt{\alpha_{t-1}}\mathbf{z}_0^m + \sqrt{\alpha_{t-1}}\mathbf{z}_0^c+ \sqrt{(1-\alpha_{t-1})} \epsilon_{t-1} \\
    &\mathbf{z}_t^m = \sqrt{\alpha_t}\mathbf{z}_0^m + \sqrt{\alpha_t}\mathbf{z}_0^c+ \sqrt{(1-\alpha_t)} \epsilon_t
\end{aligned}
\end{equation}
if the alternative inference process is non-Markovian, what we need to do is identify the relationships of changes between adjacent sampling processes.
Therefore, we can let 
\begin{equation}
    \mathbf{z}_{t-1}^m = a\mathbf{z}_{t}^m +b\mathbf{z}_{0}^m+ b\mathbf{z}_0^c+ d\epsilon_{t-1}
\end{equation}
so,
\begin{equation}
    \begin{aligned}
        &\mathbf{z}_{t-1}^m = a\mathbf{z}_{t}^m +b\mathbf{z}_{0}^m+ b\mathbf{z}_0^c+ d\epsilon_{t-1}\\
&=a(\sqrt{\alpha_t}\mathbf{z}_0^m + \sqrt{\alpha_t}\mathbf{z}_0^c+ \sqrt{(1-\alpha_t)} \epsilon_t)+b\mathbf{z}_{0}^m+ b\mathbf{z}_0^c+ d\epsilon_{t-1}\\
&=(a\sqrt{\alpha_t}+b)\mathbf{z}_{0}^m+a\sqrt{\alpha_t}+b)\mathbf{z}_0^c+a\sqrt{(1-\alpha_t)}\epsilon_t+d\epsilon_{t-1}\\
&=(a\sqrt{\alpha_t}+b)\mathbf{z}_{0}^m+a\sqrt{\alpha_t}+b)\mathbf{z}_0^c+\sqrt{(a^2(1-\alpha_t) +d^2)}\epsilon_{t-1}
    \end{aligned}
\end{equation}
because of 
\begin{equation}
    \begin{aligned}
        &\sqrt{\alpha_{t-1}} = a\sqrt{\alpha_t}+b \\
        &\sqrt{(1-\alpha_{t-1})} = \sqrt{a^2(1-\alpha_t) +d^2}
    \end{aligned}
\end{equation}
we can get 
\begin{equation}
    \begin{aligned}
        &a=\sqrt{\frac{1-\alpha_{t-1} -d^2}{1-\alpha_t}} \\
        &b=\sqrt{\alpha_{t-1}}-a\sqrt{\alpha_t}
    \end{aligned}
\end{equation}
According to Eq.\eqref{eq5}, 
\begin{equation}
    \mathbf{z}_0^m =\frac{\mathbf{z}_t^m -  \sqrt{\alpha_t}\mathbf{z}_0^c- \sqrt{(1-\alpha_t)} \epsilon_t}{\sqrt{\alpha_t}}
\end{equation}
Finally,
\begin{equation}
    \begin{aligned}
        &\mathbf{z}_{t-1}^m = a\mathbf{z}_{t}^m +b\mathbf{z}_{0}^m+ b\mathbf{z}_0^c+ d\epsilon_{t-1}\\
&=\sqrt{\frac{1-\alpha_{t-1} -d^2}{(1-\alpha_t)}}\mathbf{z}_{t}^m +(\sqrt{\alpha_{t-1}}\\
&-\sqrt{\frac{1-\alpha_{t-1}-d^2}{(1-\alpha_t)}}\sqrt{\alpha_t})\mathbf{z}_{0}^m\\
&+(\sqrt{\alpha_{t-1}}-\sqrt{\frac{1-\alpha_{t-1}-d^2}{(1-\alpha_t)}}\sqrt{\alpha_t})\mathbf{z}_0^c+d\epsilon_{t-1}\\
&=\sqrt{\frac{1-\alpha_{t-1}-d^2}{(1-\alpha_t)}}\mathbf{z}_{t}^m+(\sqrt{\alpha_{t-1}}-\sqrt{\frac{1-\alpha_{t-1}-d^2}{(1-\alpha_t)}}\sqrt{\alpha_t})\\
&\cdot \frac{\mathbf{z}_t^m-\sqrt{\alpha_t}\mathbf{z}_0^c-\sqrt{(1-\alpha_t)}\epsilon_t}{\sqrt{\alpha_t}}\\
&+(\sqrt{\alpha_{t-1}}-\sqrt{\frac{1-\alpha_{t-1}-d^2}{(1-\alpha_t)}}\sqrt{\alpha_t})\mathbf{z}_0^c+d\epsilon_{t-1}\\
    \end{aligned}
\end{equation}

\begin{equation}
    \begin{aligned}
&=\frac{\sqrt{\alpha_{t-1}}}{\sqrt{\alpha_{t}}}\mathbf{z}_t^m-(\sqrt{\alpha_{t-1}}-\sqrt{\frac{1-\alpha_{t-1}-d^2}{(1-\alpha_t)}}\sqrt{\alpha_t})\mathbf{z}_0^c-\\
&(\sqrt{\alpha_{t-1}}-\sqrt{\frac{1-\alpha_{t-1}-d^2}{(1-\alpha_t)}}\sqrt{\alpha_t})\frac{\sqrt{1-\alpha_t}}{\sqrt{\alpha_t}}\epsilon_{\theta}\\
&+(\sqrt{\alpha_{t-1}}-\sqrt{\frac{1-\alpha_{t-1}-d^2}{(1-\alpha_t)}}\sqrt{\alpha_t})\mathbf{z}_0^c+d\epsilon_{t-1}\\
&=\frac{\sqrt{\alpha_{t-1}}}{\sqrt{\alpha_{t}}}\mathbf{z}_t^m-(\sqrt{\alpha_{t-1}}-\sqrt{\frac{1-\alpha_{t-1}-d^2}{(1-\alpha_t)}}\sqrt{\alpha_t})\frac{\sqrt{1-\alpha_t}}{\sqrt{\alpha_t}}\epsilon_{\theta}\\
&+d\epsilon_{t-1}\\
&=\frac{\sqrt{\alpha_{t-1}}}{\sqrt{\alpha_{t}}}\mathbf{z}_t^m-(\sqrt{\frac{\alpha_{t-1}(1-\alpha_t)}{\alpha_t}}-\sqrt{1-\alpha_{t-1}-d^2})\epsilon_{\theta}\\
&+d\epsilon_{t-1}\\
&=\frac{\sqrt{\alpha_{t-1}}}{\sqrt{\alpha_{t}}}\mathbf{z}_t^m+(\sqrt{1-\alpha_{t-1}-d^2}-\sqrt{\frac{\alpha_{t-1}(1-\alpha_t)}{\alpha_t}})\epsilon_{\theta}\\
&+d\epsilon_{t-1}
    \end{aligned}
\end{equation}

\end{document}